\documentclass[review]{elsarticle}

\usepackage{lineno,hyperref}
\usepackage[english]{babel}
\usepackage{amssymb}% http://ctan.org/pkg/amssymb
\usepackage{pifont}% http://ctan.org/pkg/pifont
\usepackage{graphicx}
\newcommand{\cmark}{\ding{51}}%

\usepackage{pdflscape}

\usepackage{lineno,hyperref}
\usepackage{amsmath}
\usepackage{multicol}
\usepackage{verbatim}
\usepackage{float}
\restylefloat{table}

\modulolinenumbers[5]

\journal{Artificial Intelligence Review}

\begin{document}

\begin{frontmatter}

\title{Over a Decade of Social Opinion Mining: A Systematic Review}
%\tnotetext[mytitlenote]{Fully documented templates are available in the elsarticle package on \href{http://www.ctan.org/tex-archive/macros/latex/contrib/elsarticle}{CTAN}.}

\author{Keith Cortis}
\address{ADAPT Centre, \\
         Dublin City University, \\
         Ireland \\
         }
\address{keith.cortis@adaptcentre.ie}

\author{Brian Davis}
\address{ADAPT Centre, \\
         Dublin City University, \\
         Ireland \\
         }
\address{brian.davis@adaptcentre.ie}

\begin{abstract}
Social media popularity and importance is on the increase due to people using it for various types of social interaction across multiple channels. This systematic review focuses on the evolving research area of Social Opinion Mining, tasked with the identification of multiple opinion dimensions, such as subjectivity, sentiment polarity, emotion, affect, sarcasm and irony, from user-generated content represented across multiple social media platforms and in various media formats, like text, image, video and audio. Through Social Opinion Mining, natural language can be understood in terms of the different opinion dimensions, as expressed by humans. This contributes towards the evolution of Artificial Intelligence which in turn helps the advancement of several real-world use cases, such as customer service and decision making. 
A thorough systematic review was carried out on Social Opinion Mining research which totals 485 published studies and spans a period of twelve years between 2007 and 2018.   
The in-depth analysis focuses on the social media platforms, techniques, social datasets, language, modality, tools and technologies, and other aspects derived. 
Social Opinion Mining can be utilised in many application areas, ranging from marketing, advertising and sales for product/service management, and in multiple domains and industries, such as politics, technology, finance, healthcare, sports and government. The latest developments in Social Opinion Mining beyond 2018 are also presented together with future research directions, with the aim of leaving a wider academic and societal impact in several real-world applications.
\end{abstract}


\end{frontmatter}

%\linenumbers

\section{Introduction}
\label{sec_intro}

Social media is increasing in popularity and also in its importance.  This is principally due to the large number of people who make use of different social media platforms for various types of social interaction. Kaplan and Haenlein define social media as ``a group of Internet-based applications that build on the ideological and technological foundations of Web 2.0, which allows the creation and exchange of user generated content" \cite{Kaplan2010}. This definition fully reflects that social media platforms are essential for online users to submit their views and also read the ones posted by other people about various aspects and/or entities, such as opinions about a political party they are supporting in an upcoming election, recommendations of products to buy, restaurants to eat in and holiday destinations to visit. 
In particular, people's social opinions as expressed through various social media platforms can be beneficial in several domains, used in several applications and applied in real-life scenarios.  
Therefore, mining of people's opinions, which are usually expressed in various media formats, such as textual (e.g., online posts, newswires), visual (e.g., images, videos) and audio, is a valuable business asset that can be utilised in many ways ranging from marketing strategies to product or service improvement. However as Ravi and Ravi indicate in \cite{Ravi2015}, dealing with unstructured data, such as video, speech, audio and text, creates crucial research challenges. 

This research area is evolving due to the rise of social media platforms, where several work already exists on the analysis of sentiment polarity. Moreover, researchers can gauge widespread opinions from user-generated content and better model and understand human beliefs and their behaviour.
Opinion Mining is regarded as a challenging Natural Language Processing (NLP) problem, in particular for social data obtained from social media platforms, such as Twitter\footnote{http://www.twitter.com/}, and also for transcribed text. Standard linguistic processing tools were built and developed on newswires and review-related data due to such data following more strict grammar rules. These differences should be taken in consideration when performing any kind of analysis \cite{Balazs2016}. Therefore, social data is difficult to analyse due to the short length in text, the non-standard abbreviations used, the high sparse representation of terms and difficulties in finding out the synonyms and any other relations between terms, emoticons and hashtags used, lack of punctuations, use of informal text, slang, non-standard shortcuts and word concatenations. Hence, typical NLP solutions are not likely to work well for Opinion Mining.

%define what opinion mining is
\textbf{Opinion Mining} --presently a very popular field of study-- is defined by Liu and Zhang as ``the computational study of people's opinions, appraisals, attitudes, and emotions toward entities, individuals, issues, events, topics and their attributes" \cite{Liu2012}. 
%define what social is
\textbf{Social} is defined by the Merriam-Webster Online dictionary\footnote{http://www.merriam-webster.com/}
 as ``of or relating to human society, the interaction of the individual and the group, or the welfare of human beings as members of society". 

In light of this, we define \textbf{Social Opinion Mining} (SOM) as ``the study of user-generated content by a selective portion of society be it an individual or group, specifically those who express their opinion about a particular entity, individual, issue, event and/or topic via social media interaction".

Therefore, the research area of SOM is tasked with the identification of several dimensions of opinion, such as sentiment polarity, emotion, sarcasm, irony and mood, from social data which is represented in structured, semi-structured and/or unstructured data formats. Information fusion is the field tasked with researching about efficient methods for automatically or semi-automatically transforming information from different sources into a single coherent representation, which can be used to guide the fusion process. This is important due to the diversity in data in terms of content, format and volume \cite{Balazs2016}. Sections \ref{sec_om_som} and \ref{sec_issues} provide information about SOM and its challenges. 

In addition, SOM is generally very personal to the individual responsible for expressing an opinion about an object or set of objects, thus making it user-oriented from an opinion point-of-view, e.g., a social post about an event on Twitter, a professional post about a job opening on LinkedIn\footnote{https://www.linkedin.com/} or a review about a hotel on TripAdvisor\footnote{https://www.tripadvisor.com/}. 

Our SOM research focuses on microposts -- i.e. information published on the Web that is small in size and requires minimal effort to publish \cite{Microposts2016} -- that are expressed by individuals on a microblogging service, such as Sina Weibo\footnote{http://www.weibo.com} or Twitter and/or a social network service that has its own microblogging feature, such as Facebook\footnote{https://www.facebook.com/} and LinkedIn. 

\subsection{Opinion Mining vs. Social Opinion Mining}
\label{sec_om_som} 

In 2008, Pang and Lee had already identified the relevance between the field of ``social media monitoring and analysis" and the body of work reviewed in \cite{Pang2008}, which deals with the computational treatment of opinion, sentiment and subjectivity in text. This work is nowadays known as \textbf{opinion mining}, \textbf{sentiment analysis}, and/or \textbf{subjectivity analysis} \cite{Pang2008}. Other phrases, such as \textbf{review mining} and \textbf{appraisal extraction} have also been used in the same context, whereas some connections have been found to \textbf{affective computing} (where one of its goals is to enable computers in recognising and expressing emotions) \cite{Pang2008}. Merriam-Webster's Online Dictionary defines that the terms\footnote{http://www.merriam-webster.com/dictionary/opinion} ``opinion", ``view", ``belief", ``conviction", ``persuasion" and ``sentiment" mean a judgement one holds as true. This shows that the distinctions in common usage between these terms can be quite subtle.  
In light of this, three main three research areas --opinion mining, sentiment analysis and subjectivity analysis-- are all related and use multiple techniques taken from NLP, information retrieval, structured and unstructured data mining \cite{Ravi2015}. However, even though these three concepts are broadly used as synonyms, thus used interchangeably, it is worth noting that their origins differ. Some authors also consider that each concept presents a different understanding \cite{Serrano2015}, and also have different notions \cite{Tsytsarau2012}. We are in agreement with this, hence we felt that a new terminology is required to properly specify what SOM means, as defined in Section \ref{sec_intro}.  

According to Cambria et al., sentiment analysis can be considered as a very restricted NLP problem, where the polarity (negative/positive) of each sentence and/or target entities or topics needs to be understood \cite{Cambria2013}. On the other hand, Liu discusses that ``opinions are usually subjective expressions that describe people's sentiments, appraisals or feelings toward entities, events and their properties" \cite{Liu2010}. He further identifies two sub-topics of sentiment and subjectivity analysis, namely sentiment classification (or document-level sentiment classification) and subjectivity classification. SOM requires such classification methods to determine an opinion dimension, such as objectivity/subjectivity and sentiment polarity. For example, subjectivity classification is required to classify whether user-generated content, such as a product review, is objective or subjective, whereas sentiment classification is performed on subjective content to find the sentiment polarity (positive/negative) as expressed by the author of the opinionated text. In cases where the user-generated content is made up of multiple sentences, sentence-level classification needs to be performed to determine the respective opinion dimension. In addition, sentence-level classification is not suitable for compound sentences, i.e., a sentence that expresses more than one opinion. For such cases, aspect-based opinion mining needs to be performed. 

\subsection{Issues and Challenges}
\label{sec_issues}
In 2008, Pang and Lee \cite{Pang2008} had already identified that the writings of Web users can be very challenging in their own way due to numerous factors, such as the quality of written text, discourse structure and the order in which different opinions are presented. The effects of the latter factor can result in a completely opposite overall sentiment polarity, where the order effects can completely overwhelm the frequency effects. This is not the case in traditional text classification, where if a document refers to the term ``car" in a frequent manner, the document will probably somewhat be related to cars. Therefore, order dependence manifests itself in a more fine-grained level of analysis.

Liu in \cite{Liu2010} mentions that complete sentences (for reviews) are more complex than short phrases and contain a large amount of noise, thus making it more difficult to extract features for feature-based sentiment analysis. Even though we agree that with more text, comes a higher probability of spelling mistakes, etc., we tend to disagree that shorter text, such as microposts, contain less noise. 

The process of mining user-generated content posted on the Web is very intricate and challenging due to the nature of short textual content limit (e.g., tweets allowed up to 140 characters till October 2017), which at times forces a user to resort in using short words, such as acronyms and slang, to make a statement. These often lead to further issues in the text, such as misspellings, incomplete content, jargon, incorrect acronyms and/or abbreviations, emoticons and content misinterpretation \cite{Cortis2013}. Other noteworthy challenges include swear words, irony, sarcasm, negatives, conditional statements, grammatical mistakes, use of multiple languages, incorrect language syntax, syntactically inconsistent words, and different discourse structures. In fact, when informal language is used in the user-generated content, the grammar and lexicon varies from the standard language normally used \cite{Dashtipour2016}. Moreover, user-generated text exhibits more language variation due to it being less grammatical than longer posts, where the aforementioned use of emoticons, abbreviations together with hashtags and inconsistent capitalisation, can form an important part of the meaning \cite{Maynard2012}. In \cite{Maynard2012}, Maynard et al. also points out that microposts are in some sense the most challenging type of text for text mining tools especially for opinion mining, since they do not contain a lot of contextual information and assume much implicit knowledge. Another issue is ambiguity, since microposts such as tweets do not follow a conversation thread. Therefore, this isolation from other tweets makes it more difficult to make use of coreference information unlike in blog posts and comments. Due to the short textual content, features can also be sparse to find and use, in terms of text representation \cite{WangMin2014}. In addition, the majority of microposts usually contain information about a single topic due to the length limitation, which is not the case in traditional blogs, where they contain information on more than one topic given that they do not face the same length limitations \cite{Giachanou2016c}.

Big data challenges, such as handling and processing large volumes of streaming data, are also encountered when analysing social data \cite{Bravo2014}. 
Limited availability of labelled data and dealing with the evolving nature of social streams usually results in the target concept changing which would require the learning models to be constantly updated \cite{Guerra2014}. 

In light of the above, social networking services bring several issues and challenges with them and the way in how content is generated by their users. Therefore, several Information Extraction tasks, such as Named Entity Recognition (NER) and Coreference Resolution, might be required to carry out multi-dimensional SOM. In fact, several shared evaluation tasks are being organised to try and reach a standard mechanism towards performing IE tasks on noisy text which is very common in user-generated social media content. As already discussed in detail above, such tasks are much harder to solve when they are applied on micro-text like microposts \cite{Ravi2015}. This problem presents serious challenges on several levels, such as performance. Examples of such tasks are ``Named Entity Recognition in Twitter"\footnote{http://noisy-text.github.io/2016/ner-shared-task.html}. 

In terms of content, social media-based studies present only analysis and results from a selective portion of society, since not everyone uses social media. Moreover, several cross-cultural differences and factors determine the social media usage in each country and hence the results of such studies. For example for the Political domain, these services are used predominantly by young and politically active individuals or by ones with strong political views. This could be easily reflected in the Brexit results, where the majority of younger generation (age 18-44) voted to remain in the European Union as opposed to people over age 45. Such a result falls in line with the latest United Kingdom social media statistics, such as for Twitter, where 72\% of the users are between the age of 15-44, whilst for Facebook the most popular age group is 25-34 (26\% of users) \cite{Hurlimann2016}. However, results of similar studies in other cultures and languages might differ due to different use of social words to reflect a general opinion, sentiment polarity and/or emotion \cite{Lin2018}.  

\subsection{Systematic Review}
\label{sec_systematic}

In light of the above, it is noteworthy that no systematic review within this newly defined domain exists even though there are several good survey papers \cite{Liu2012, Tsytsarau2012, Medhat2014, Ravi2015}. The research paper by Bukhari et al. \cite{Bukhari2016} is closest to a systematic review in this domain, whereby the authors performed a search over the ScienceDirect and SpringerLink electronic libraries for the ``sentiment analysis", ``sentiment analysis models", ``sentiment analysis of microblogs" terms. As a result, we felt that the SOM domain well and truly deserves a thorough systematic review that captures all of the relevant research conducted over the last decade. This review also identifies the current literature gaps within this popular and constantly evolving research domain.

The structure of this comprehensive systematic review is as follows: Section \ref{sec_research_method} presents the research method adopted to carry out this review, followed by Section \ref{sec_review_analysis} which provides a thorough review analysis of the main aspects derived from the analysed studies. This is followed by Section \ref{sec_social_opinion_mining-forms} which focuses on the different dimensions of social opinions as derived from the analysed studies, and Section \ref{sec_social_opinion_mining-areas} which presents the application areas where SOM is being used. Lastly, Section \ref{sec_conc_rem} discusses the the latest developments for SOM (beyond the period covered by the systematic review) and future research directions as identified by the authors.

\section{Research Method}
\label{sec_research_method}

This survey paper about SOM adopts a systematic literature review process. This empirical research process was based on the guidelines and procedures proposed by \cite{Kitchenham2004, Brereton2007, Dyba2007, Attard2015} which were focused on the software engineering domain. The systematic review process although more time consuming is reproducible, minimising bias and maximising internal and external validity. The procedure undertaken was structured as follows and is explained in detail within the sub-sections below:

\begin{enumerate}
\item Specification of research questions;
\item Generation of search strategy which includes the identification of electronic sources (libraries) and selection of relevant search terms;
\item Application of the relevant search;
\item Choice of primary studies via the utilisation of inclusion and exclusion criteria on the obtained results;
\item Extraction of required data from primary studies;
\item Synthesis of data. 
\end{enumerate}

\subsection{Research Questions}
\label{ssec_research_questions}

A systematic literature review is usually characterised by an appropriate generic ``research question, topic area, or phenomenon of interest" \cite{Kitchenham2004}. This question can be expanded into a set of sub-questions that are more clearly defined, whereby all available research relevant to these sub-questions are identified, evaluated and interpreted. 

The goal of this systematic review is to identify, analyse and evaluate current opinion mining solutions that make use of social data (data extracted from social media platforms). In light of this, the following generic research question is defined: 
\begin{itemize}
\item \textbf{What are the existing opinion mining approaches which make use of user-generated content obtained from social media platforms?}
\end{itemize}

The following are specific sub-questions that the generic question above can be sub-divided into:
\begin{enumerate}
\item What are the existing approaches that make use of social data for opinion mining and how can they be classified\footnote{Classification in this context refers to the dimension of opinion mining being conducted, e.g., subjectivity detection, sentiment analysis, sarcasm detection, emotion analysis, etc.}? 
\item What are the different dimensions/types of social opinion mining? 
\item What are the challenges faced when performing opinion mining on social data?
\item What techniques, datasets, tools/technologies and resources are used in the current solutions?
\item What are the application areas of social opinion mining?
\end{enumerate}

\subsection{Search Strategy}
\label{ssec_search_strategy}

The search strategy for this systematic review is primarily directed via the use of published papers which consist of journals, conference/workshop proceedings, or technical reports. The following electronic libraries were identified for use, due to their wide coverage of relevant publications within our domain: ACM Digital Library\footnote{https://dl.acm.org/}, IEEE Xplore Digital Library\footnote{http://ieeexplore.ieee.org/Xplore/home.jsp}, ScienceDirect\footnote{https://www.sciencedirect.com/}, and SpringerLink\footnote{https://link.springer.com/}.

The first three electronic libraries listed were used by three out of the four systematic reviews that our research process was based on (and which made use of a digital source), whereas SpringerLink is one of the most popular sources for publishing work in this domain (as will be seen in Section \ref{ssec_study_selection} below). Moreover, three other electronic libraries were considered for use, two --Web of Science\footnote{https://webofknowledge.com/} and Ei Compendex\footnote{https://www.elsevier.com/solutions/engineering-village/content/compendex}-- which the host university did not have access to and Google Scholar\footnote{http://scholar.google.com/} which was not included, since content is obtained from the electronic libraries listed above (and more), thus making the process redundant.

The relevant search terms were identified for answering the research questions defined in Section \ref{ssec_research_questions}. In addition, these questions were also used to perform some trial searches before the following list of relevant search terms was determined:

\begin{multicols}{2}
    \begin{enumerate}
        \item ``Social opinion mining";
		\item ``Social sentiment analysis";
		\item ``Opinion mining social media";
		\item ``Sentiment analysis social media";
		\item ``Microblog opinion mining"; 
		\item ``Microblog sentiment analysis";
		\item ``Social network sentiment"; 
		\item ``Social network opinion";
		\item ``Social data sentiment analysis";
		\item ``Social data opinion mining"; 
		\item ``Twitter sentiment analysis";
		\item ``Twitter opinion mining"; 
		\item ``Social data analysis".
    \end{enumerate}
\end{multicols}

The following are important justifications behind the search terms selected above:
\begin{itemize}
	\item ``opinion mining" and ``sentiment analysis": are both included due to the fact that these key terms are used interchangeably to denote the same field of study \cite{Pang2008, Cambria2013}, even though their origins differ and hence do not refer to the same concept \cite{Serrano2015};
	\item ``microblog", ``social network" and ``Twitter": the majority of the opinion mining and/or sentiment analysis research and development efforts target these two kinds of social media platforms, in particular the Twitter microblogging service.
\end{itemize}

\subsection{Search Application}
\label{ssec_search_application}

The ``OR" Boolean operator was chosen to formulate the search string. The search terms were all linked using this operator, making the search query simple and easy to use across multiple electronic libraries. Therefore, a publication only had to include any one of the search terms to be retrieved \cite{Attard2015}. In addition, this operator is more suitable for the defined search terms given that this study is not a general one e.g., about opinion mining in general, but is focused about opinion mining in a social context. Construction of the correct search string (and terms) is very important, since this eliminates noise (i.e. false positives) as much as possible and at the same time still retrieves potential relevant publication which increases recall. 

Several other factors had to be taken in consideration during the application of search terms on the electronic libraries. The following is a list of factors relevant to our study, identified in \cite{Brereton2007} and verified during our search application process:

\begin{itemize}
\item Electronic library search engines have different underlying models, thus not always provide required support for systematic searching;
\item Same set of search terms cannot be used for multiple engines e.g., complex logical combination not supported by the ACM Digital Library but is by the IEEE Xplore Digital Library;
\item Boolean search string is dependent on the order of terms, independent of brackets;
\item Inconsistencies in the order or relevance in search results (e.g., IEEE Xplore Digital Library results are sorted in order of relevance);
\item Certain electronic libraries treat multiple words as a Boolean term and look for instances of all the words together (e.g., ``social opinion mining"). In this case, the use of the ``AND" Boolean operator (e.g., ``social AND opinion AND mining") looks for all of the words but not necessary together. 
\end{itemize}

On the above, in our case it was very important to select a search strategy that is more appropriate to the review's research question which could be applied to the selected electronic libraries.

When applying the relevant search on top of the search strategy defined in Section \ref{ssec_search_strategy}, another important element was to identify appropriate metadata fields upon which the search string can be executed. Table \ref{table:metadata-fields} presents the ones applied in our study.

%\vspace*{-4mm}

\begin{table} [h]
    \scriptsize
	\begin{center}
    	\begin{tabular}{ | c | c | c | c | c | }
    	\hline
     	\textbf{Metadata field} & \textbf{ACM} & \textbf{IEEE Xplore}  & \textbf{ScienceDirect} & \textbf{SpringerLink} \\ \hline
     	title          &  \checkmark & \checkmark & \checkmark     & \checkmark   \\ \hline  
     	abstract       &  \checkmark & \checkmark & \checkmark     & \checkmark   \\ \hline  
     	keywords       &  \checkmark & \checkmark & \checkmark     &         \\ \hline 
    	\end{tabular}
	\end{center}
	\caption{Metadata fields used in search application}
	\label{table:metadata-fields}
\end{table}

%\vspace*{-8mm}

Applying the search on the title metadata field alone would result in several missed and/or incorrect results. Therefore, using the abstract and/or keywords in the search is very important to reduce the number of irrelevant results. In addition, this ensures that significant publications that lack any of the relevant search terms within their title are returned.

A separate search method was applied for each electronic library, since they all offer different functionalities and have different underlying models. Each method is detailed below:
\begin{itemize}
\item ACM: Separate searches for each metadata field were conducted and results were merged (duplicates removed). Reason being that the metadata field search functionality ``ANDs" all metadata fields, whereas manual edition of the search query does not work well when amended. 
\item IEEE: Separate searches for each metadata field were conducted and results were merged (duplicates removed).
\item ScienceDirect: One search that takes in consideration all the chosen metadata fields.
\item SpringerLink: By entering a search term or phrase, a search is conducted over the title, abstract and full-text (including authors, affiliations and references) of every article and book chapter. This was noted in the large amount of returned papers (as will be discussed in the next sub-section), which results in a high amount of false positives (and possibly a higher recall).  
\end{itemize}

\subsection{Study Selection}
\label{ssec_study_selection}

A manual study selection was performed on the primary studies obtained from the search application defined in Section \ref{ssec_search_application}. This is required to eliminate any studies that might be irrelevant even though the search terms appear in either of the metadata fields defined in Table \ref{table:metadata-fields} above. Therefore, inclusion and exclusion criteria (listed below) were defined. 

Published papers that meet any of the following inclusion criteria are chosen as primary studies:
\begin{itemize}
\item I1. A study that targeted at least one social networking service and/or utilised a social dataset besides other social media services, such as blogs, chats and wikis. Please note that only work performed on social data from social networking services is taken in consideration for the purposes of this review;
\item  I2. A study published from the year 2007 onwards. This year was chosen, since the mid-2000s saw the evolution of several social networking services, in particular Facebook's growth (2007), which currently contains the highest monthly active users;  
\item I3. A study published in the English language.
\end{itemize}

Published papers that satisfy any of the exclusion criteria from the following list, are removed from the systematic review:
\begin{itemize}
\item E1. A study published before 2007;
\item E2. A study that does not focus on performing any sort of opinion mining on social media services, even though it mentions some of the search terms;
\item E3. A study that focuses on opinion mining or sentiment analysis in general i.e. no reference in a social context;
\item E4. A study that is only focused on social data sources obtained from online forums, communities, blogs, chats, social news websites (e.g., Slashdot\footnote{https://slashdot.org/}), review websites (e.g., IMDb\footnote{https://www.imdb.com/});
%, Epinions);
\item E5. A study that consists of either a paper's front cover and/or title page. 
\end{itemize}

Selection of the primary studies for this systematic review was carried out in 2019. Therefore, studies indexed or published from 2019 onwards, are not included in this review.

%\vspace*{-4mm}

\begin{table} [h!]
    \scriptsize
	\begin{center}
    	\begin{tabular}{ | c | c | c | c | c | }
    	\hline
     	\textbf{Primary studies} & \textbf{ACM} & \textbf{IEEE Xplore}  & \textbf{ScienceDirect} & \textbf{SpringerLink} \\ \hline
     	Search application    & 106 & 242  & 57 & 456   \\ \hline  
     	False positives       & 39  & 83   & 17 & 262   \\ \hline       	
     	Study selection       & 67  & 159  & 40 & 194   \\ \hline 
     	No full paper access  &  0  &  0   & 5  & 4     \\ \hline 
     	Full paper access     & 67  & 159  & 35 & 190   \\ \hline
     	
     	Total                 & \multicolumn{4}{|c|}{\text{451}} \\ \hline
    	\end{tabular}
	\end{center}
	\caption{Primary studies selection procedure from the electronic libraries}
	\label{table:primary-studies}
\end{table}

%\vspace*{-10mm} 

Table \ref{table:primary-studies} shows the results for each electronic library at each step of the procedure used for selecting the final set of primary studies. The results included one proceedings, which was resolved by including all the published papers within the track relevant to this study\footnote{Proceedings returned from IEEE Xplore: \url{http://ieeexplore.ieee.org/stamp/stamp.jsp?reload=true&arnumber=7344780} where the ``Emotion and Sentiment in Intelligent Systems and Big Social Data Analysis (SentISData)" track was relevant for this study.}, since the other papers were not relevant thus not included in the initial results. The search application phase resulted in a total of 861 published papers. False positives, which consist of duplicate papers and papers that meet any of the exclusion criteria were removed. This was done through a manual study selection which was performed on all the metadata fields considered i.e. the title, abstract and keywords. In cases where we were still unclear of whether a published paper is valid or not, we went through the full text. This study selection operation left us with 460 published papers, where the number of false positives totalled 401. Out of the final study selection published papers, we did not have full access to 9 published papers, thus reducing the total primary studies to 451. 

In addition to the primary studies selected from the electronic libraries, we added a set of relevant studies --34 published papers (excluding survey papers)-- for completeness sake which were either published in reputable venues within the Opinion Mining community or were highly cited.
Therefore, the final set of primary studies totals 485 published papers.        
 
\subsection{Extraction of data}
\label{ssec_extraction_data}

\subsubsection{Overall}
\label{sssec_overall_extraction}

The main objective of this study is to conduct a systematic analysis of the current literature in the field of SOM. Each published paper in this review was analysed in terms of the following information/parameters: social media platforms, techniques and approaches, social datasets, language, modality, tools and technologies, (other) NLP tasks, application areas and opinion mining dimensions. It is important to note that this information was manually extracted from each published paper. In the sub-sections below we discuss the overall statistics about the relevant primary studies that resulted from the study selection phase of this systematic review. 

\subsubsection{Study Selection - Electronic Libraries}

\begin{figure}[h!]
	\centering    
    \includegraphics[width=1\textwidth]{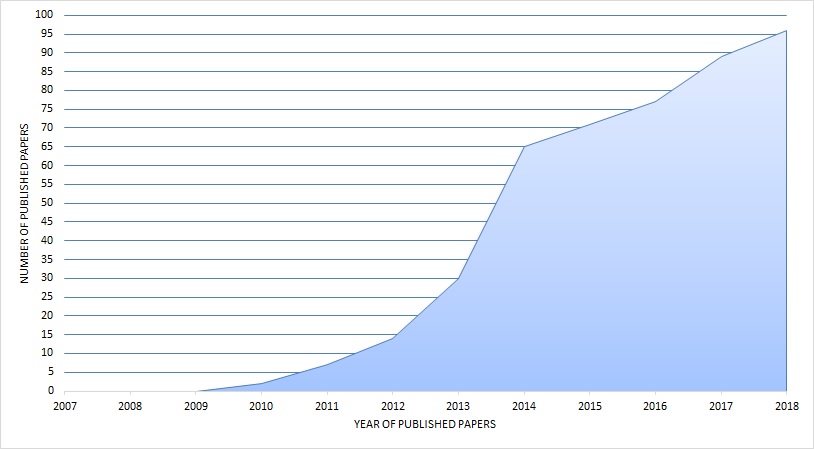}
	\caption{Primary Studies by Year}
	\label{fig:primary-studies-yearly}
\end{figure}

Figure \ref{fig:primary-studies-yearly} shows that the first three years of this evaluation period, i.e., 2007-2009, did not return any relevant literature. It is important to note that 2006 and 2007 was the period when opinion mining emerged in Web applications and weblogs within multiple domains, such as politics and marketing \cite{Pang2008}. However, 2010 -- which year coincides with the introduction of various social media platforms and the major increase in Facebook and Twitter usage\footnote{https://www.techinasia.com/social-media-timeline-2010} -- resulted in the first relevant literature, which figures kept increasing in the following years. Please note that the final year in evaluation, that is 2018, contains literature that was published or indexed till the 31st December 2018. From the twelve full years evaluated, 2018 produced the highest number of relevant literature. This shows the importance of opinion mining on social data, and therefore the continuous increase in social media usage and popularity, in particular social networking services. Moreover, SOM solutions are on the increase for various real world applications.   

%\vspace*{-4mm}

%\vspace*{-6mm}

\subsubsection{Study Selection - Additional Set}

The additional set of studies included in this systematic review, were published in the period between the year 2009 and 2014. These ranged from various publishers, namely the four selected for this study (ACM, IEEE Xplore, ScieneDirect and SpringerLink) and other popular ones, such as Association for the Advancement of Artificial Intelligence (AAAI)\footnote{https://aaai.org/Library/library.php}, Association for Computational Linguistics (ACL)\footnote{http://aclweb.org/anthology/} and Wiley Online Library\footnote{https://onlinelibrary.wiley.com/}.  

\subsection{Synthesis of data}
\label{ssec_synthesis_data}

The data synthesis of this detailed analysis is based on the extracted data mentioned in Section \ref{sssec_overall_extraction} above, which is discussed in the subsequent sections. 

\section{Review Analysis}
\label{sec_review_analysis}

Table \ref{table:primary-studies-categorisation} provides different high level categories of the primary studies selected for this systematic review, discussed in Section \ref{ssec_study_selection}. 

%\vspace*{-5mm}

\begin{table} [h!]
    \scriptsize
	\begin{center}
    	\begin{tabular}{ | p{3.75cm} | c | p{1cm} | p{1cm} | p{1.25cm} | p{1.75cm} | }
    	\hline
        \textbf{Categories}  & \textbf{ACM} & \textbf{IEEE Xplore}  & \textbf{Science Direct} & \textbf{Springer Link} & \textbf{Additional Set} \\ \hline
     	Study selection         & 67  & 159           & 40            & 194   		& 34 \\ \hline 
     	No full paper access    &  0  &  0	         & 5             & 4     		& 0  \\ \hline 
        Surveys                 &  2  &  5           & 3             & 8			& 0  \\ \hline
        Work can be applied/used on social data     &  1  &  0           & 0             & 1			& 0  \\ \hline
        Organised tasks         &  0  &  0           & 0             & 2			& 0  \\ \hline
    	\end{tabular}
	\end{center}
	\caption{Categories of primary studies}
	\label{table:primary-studies-categorisation}
\end{table}

%\vspace*{-6mm}

It must be noted that not all the published papers were considered in the analysis conducted. Therefore, this table is referenced in all of the different aspects of the data synthesised, as presented below. 
It presents the primary studies returned from each electronic library and the additional ones, together with the ones that do not have full access, survey papers, papers which present work that can be applied/used on social data, and papers originating from organised tasks within the domain. 

The in-depth analysis, which focused on the social media platforms, techniques, social datasets, language, modality, tools and technologies, NLP tasks and other aspects used across the published papers, is presented in Sections \ref{ssec_social_media_platforms}-\ref{ssec_nlp}.

\subsection{Social Media Platforms}
\label{ssec_social_media_platforms}

Social data refers to online data generated from any type of social media platform be it from microblogging, social networking, blogging, photo/video sharing and crowdsourcing. 
Given that this systematic survey focuses on opinion mining approaches that make use of social networking and microblogging services, we identify the social media platforms used in the studies within this review.

In total, 469 studies were evaluated with 66 from ACM, 155 from IEEE Xplore, 32 from ScienceDirect, 182 from SpringerLink and 34 additional ones. Papers which did not provide full access were excluded. Note that 4 survey papers -- 2 from ACM \cite{Giachanou2016a, Zimbra2018}, 1 from IEEE Xplore \cite{Wagh2018}, 1 from SpringerLink \cite{AbdullahMalak2017-2} -- and 2 SpringerLink organised/shared task papers \cite{Loukachevitch2015, Patra2015} were included, since the former papers focus on Twitter Sentiment Analysis methods whereas the latter papers focus on Sentiment Analysis of tweets (therefore the target social media platform of all evaluated papers is clear in both cases). None of the other 14 survey papers \cite{Rajalakshmi2017, Yenkar2018, Abdelhameed2017, Rathan2017, LiuSam2018, ZhangWenping2018, Ravi2015, Nassirtoussi2014, Beigi2016, Lo2017, Ji2016, Batrinca2015, Li2014b, Lin2014} have been included, since various social media platforms were used in the respective studies evaluated. In addition, 2 papers that presented a general approach which can be applied/used on social data (i.e., not on any source) \cite{Min2013, ElHaddaoui2018} have also not been included.

\begin{figure}[h!]
	\centering    
    \includegraphics[width=1\textwidth]{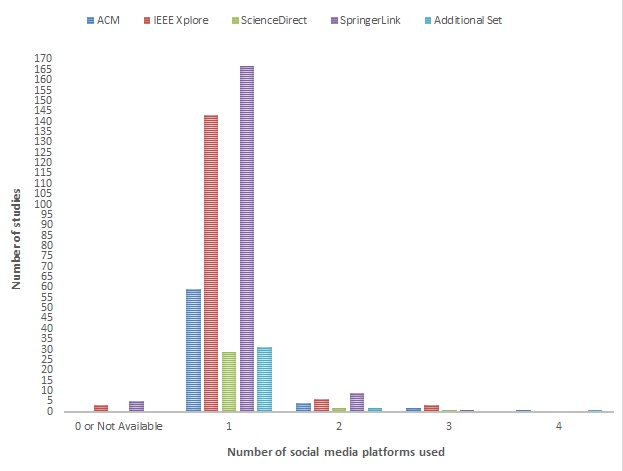}
	\caption{Number of social media platforms used in each study}
	\label{fig:social-media-platforms-per-study}
\end{figure}

Out of these studies, 429 made use of 1 social media platform, whereas 32 made use of 2 to 4 social media platforms, as can be seen in Figure \ref{fig:social-media-platforms-per-study}. 

%\vspace*{-4mm}

%\vspace*{-4mm}

With respect to social media platforms, in total 504 were used across all of the studies. These span over the following 18 different ones, which are also listed in Table \ref{table:social-media-platforms-in-studies}: 
\begin{enumerate}
	\item Twitter%\footnote{https://twitter.com/}
	: a microblogging platform that allows publishing of short text updates (``microposts");
	\item Sina Weibo%\footnote{https://weibo.com/}
	: a Chinese microblogging platform that is like a hybrid of Twitter and Facebook; 
	\item Facebook%\footnote{https://www.facebook.com/}
	: a social networking platform that allows users to connect and share content with family and friends online;
	\item YouTube\footnote{https://www.youtube.com/}: a video sharing platform;
	\item Tencent Weibo\footnote{http://t.qq.com/}: a Chinese microblogging platform; 
	\item TripAdvisor%\footnote{https://www.tripadvisor.com/}
	: a travel platform that allows people to post their reviews about hotels, restaurants and other travel-related content, besides offering accommodation bookings; 
	\item Instagram\footnote{https://www.instagram.com}: a platform for sharing photos and videos from a smartphone;
	\item Flickr\footnote{https://www.flickr.com/}: an image- and video-hosting platform that is popular for sharing personal photos; 
	\item Myspace\footnote{https://myspace.com/}: a social networking platform for musicians and bands to show and share their talent and connect with fans;
	\item Digg\footnote{http://digg.com/}: a social bookmarking and news aggregation platform that selects stories to the specific audience; 
	\item Foursquare\footnote{https://foursquare.com/}: formerly a location-based service and nowadays a local search and discovery service mobile application known as Foursquare City Guide;
	\item Stocktwits\footnote{https://stocktwits.com/}: a social networking platform for investors and traders to connect with each other;
	\item LinkedIn\footnote{https://www.linkedin.com/}: a professional networking platform that allows users to communicate and share updates with colleagues and potential clients, job searching and recruitment;
	\item Plurk\footnote{https://www.plurk.com}: a social networking and microblogging platform; 
	\item Weixin\footnote{https://weixin.qq.com/}: a Chinese multi-purpose messaging and social media app developed by Tencent;
	\item PatientsLikeMe\footnote{https://www.patientslikeme.com/}: a health information sharing platform for patients;
	\item Apontador\footnote{https://www.apontador.com.br/}: a Brazilian platform that allows users to share their opinions and photos on social networks and also book hotels and restaurants;
	\item Google+\footnote{https://plus.google.com/}: formerly a social networking platform (shut down in April 2019) that included features such as posting photos and status updates, group different relationship types into Circles, organise events and location tagging.
\end{enumerate}

\begin{table} [h!]
    \scriptsize
	\begin{center}
    	\begin{tabular}{ | p{2cm} | p{1cm} | p{1cm} | p{1cm} | p{1.25cm} | p{1.5cm} | p{1cm} | }
    	\hline
        \textbf{Social Media Platform}  & \textbf{ACM} & \textbf{IEEE Xplore}  & \textbf{Science Direct} & \textbf{Springer Link} & \textbf{Additional Set}  & \textbf{Total} \\ \hline
      Twitter    	& 53 	& 130	& 25	& 136	& 27	& 371  \\ \hline 
      Sina Weibo   	& 4  	& 13	& 1		& 26 	& 2   	& 46   \\ \hline
      Facebook     	& 4  	& 10	& 3		& 10	& 3     & 30   \\ \hline 
      YouTube      	& 7   	& 1		& 2		& 1		& 1  	& 12   \\ \hline 
      Tencent Weibo	& 0  	& 1		& 1		& 5		& 1  	& 8	   \\ \hline
      TripAdvisor  	& 0   	& 1		& 2		& 4		& 0		& 7	   \\ \hline       
      Instagram    	& 2   	& 3		& 0		& 1		& 0		& 6	   \\ \hline 
      Flickr       	& 0   	& 2		& 0		& 3		& 0  	& 5	   \\ \hline
      Myspace      	& 2  	& 0		& 0		& 0		& 3  	& 5    \\ \hline 
      Digg        	& 2   	& 0		& 0		& 0		& 1  	& 3    \\ \hline 
      Foursquare   	& 2   	& 0		& 0		& 1		& 0  	& 3    \\ \hline
      Stocktwits   	& 0  	& 1		& 1		& 0		& 0 	& 2    \\ \hline         
      LinkedIn     	& 1   	& 0		& 0		& 0		& 0  	& 1	   \\ \hline 
      Plurk        	& 0  	& 0		& 0		& 1		& 0  	& 1	   \\ \hline
      Weixin      	& 0   	& 0		& 1		& 0		& 0  	& 1	   \\ \hline 
     PatientsLikeMe	& 0  	& 1		& 0		& 0		& 0  	& 1    \\ \hline 
      Apontador    	& 0   	& 0		& 0		& 0		& 1		& 1    \\ \hline
      Google+     	& 0   	& 1		& 0		& 0		& 0  	& 1    \\ \hline           	 
    	\end{tabular}
	\end{center}
	\caption{Social media platforms used in the studies}
	\label{table:social-media-platforms-in-studies}
\end{table}

Overall, Twitter was the most popular with 371 opinion mining studies making use of it, followed by Sina Weibo with 46 and Facebook with 30. Other popular platforms such as YouTube (12), Tencent Weibo (8), TripAdvisor (7), Instagram (6) and Flickr (5) were also used in a few studies. These results show the importance and popularity of microblogging platforms, such as Twitter and Sina Weibo, which are also very frequently used for research and development purposes in this domain. Such microblogging platforms provide researchers the possibility of using an Application Programming Interface (API) to access social data, which plays a crucial role in selecting them for their studies. On the other hand, data retrieval from other social media platforms such as Facebook, is becoming more challenging due to ethical concerns. For example, Facebook access to the Public Feed API\footnote{\url{https://developers.facebook.com/docs/public_feed/}} is restricted and users cannot apply for it. 

\subsection{Techniques}
\label{ssec_techniques_approaches}

For this analysis, 465 studies were evaluated: 65 from ACM, 154 from IEEE Xplore, 32 from ScienceDirect, 180 from SpringerLink and 34 additional ones. Studies excluded are the ones with no full access, surveys, and organised task papers. The main aim was to identify the technique/s used for the opinion mining process on social data. Therefore, they were categorised under the following approaches: Lexicon (Lx), Machine Learning (ML), Deep Learning (DL), Statistical (St), Probabilistic (Pr), Fuzziness (Fz), Rule (Rl), Graph (Gr), Ontology (On), Hybrid (Hy) --a combination of more than one technique, Manual (Mn) and Other (Ot). Table \ref{table:om-approaches} provides the yearly statistics for all the respective approaches adopted.

\begin{table} [h!]
	\scriptsize
	\begin{center}
    	\begin{tabular}{ | c | c | c | c | c | c | c | c | c | c | c | c | c |}
    	\hline
     	\textbf{Year} & \textbf{Lx} & \textbf{ML} & \textbf{DL} & \textbf{St} & \textbf{Pr} & \textbf{Fz} & \textbf{Rl} & \textbf{Gr} & \textbf{On} & \textbf{Hy} & \textbf{Mn} & \textbf{Ot} \\ \hline
     	2007          & 0  & 0  & 0  & 0  & 0  & 0  & 0  & 0  & 0  & 0  & 0  & 0  \\ \hline  
     	2008          & 0  & 0  & 0  & 0  & 0  & 0  & 0  & 0  & 0  & 0  & 0  & 0  \\ \hline  
     	2009          & 0  & 1  & 0  & 0  & 0  & 0  & 0  & 0  & 0  & 1  & 0  & 0  \\ \hline  
     	2010          & 2  & 2  & 0  & 0  & 0  & 0  & 0  & 0  & 0  & 3  & 0  & 0  \\ \hline  
     	2011          & 2  & 3  & 1  & 0  & 0  & 0  & 0  & 0  & 0  & 7  & 1  & 0  \\ \hline  
     	2012          & 6  & 5  & 0  & 0  & 0  & 0  & 0  & 1  & 0  & 10 & 0  & 1  \\ \hline  
     	2013          & 6  & 14 & 2  & 1  & 1  & 0  & 2  & 0  & 1  & 21 & 0  & 0  \\ \hline  
     	2014          & 14 & 20 & 2  & 1  & 3  & 1  & 1  & 0  & 1  & 41 & 0  & 3  \\ \hline   
     	2015          & 16 & 15 & 4  & 1  & 1  & 0  & 0  & 1  & 0  & 42 & 0  & 0  \\ \hline  
     	2016          & 13 & 21 & 4  & 3  & 0  & 0  & 0  & 0  & 0  & 38 & 2  & 4  \\ \hline  
     	2017          & 20 & 22 & 9  & 2  & 1  & 1  & 0  & 0  & 0  & 50 & 2  & 5  \\ \hline  
     	2018          & 17 & 18 & 13 & 1  & 0  & 0  & 1  & 2  & 0  & 69 & 1  & 4  \\ \hline  
        \textbf{Total}& 96 & 121 & 35 & 9 & 6  & 2  & 4  & 4  & 2  & 282 & 6 & 17 \\ \hline  
    	\end{tabular}
	\end{center}
	\caption{Approaches used in the studies analysed}
	\label{table:om-approaches}
\end{table}

%\vspace*{-7mm}

From the studies analysed, 88 developed and used more than 1 technique within their respective studies. These techniques include the ones originally used in their approach and/or ones used for comparison/baseline/experimentation purposes. In particular, from these 88 studies, 65 used 2 techniques each, 17 studies used 3 techniques, 4 studies used 4 techniques, and 2 studies made use of 5 techniques, which totals to 584 techniques used across all studies (including the studies that used 1 technique). 
The results show that a hybrid approach is the most popular one, with over half of the studies adopting such an approach. This is followed by Machine Learning and Lexicon techniques, which are usually chosen to perform any form of opinion mining. These results are explained in more detail in the sub-sections below. 

\subsubsection{Lexicon}
\label{sssec_lexicon_approaches}

In total 94 unique studies adopted a lexicon-based approach to perform a form of SOM, which produced a total of 96 different techniques\footnote{The authors in \cite{Rathan2018, Hagen2015} both propose two lexicon based methods, which are included in total techniques.}. The majority of the lexicons used were specifically related to opinions and are well known in this domain, whereas the others that were not can still be used for conducting opinion mining.  

%\vspace*{-4mm}

\begin{table} [h!]
	\small
	\begin{center}
    	\begin{tabular}{ | p{1.75cm} | p{1cm} | p{1cm} | p{0.65cm} | p{0.65cm} | p{0.65cm} | p{0.65cm} | p{0.65cm} | p{1.5cm} | }
    	\hline
        \textbf{Number of lexicons}  & 1  & 2  & 3 & 4 & 6 & 7 & 8 & Other / NA \\ \hline
        \textbf{Number of studies }  & 39 & 19 & 10 & 4 & 1 & 1 & 1 & 21       \\ \hline
        \textbf{References of studies} & \cite{Rathan2018, LiYujiao2018, Geetha2018, ChenYang2018, Aoudi2018, Poortvliet2018, Salari2018, Hubert2018, Ray2017, GuptaI2017, Zhang2017, Arslan2017, Hagge2017, Ozer2017, Ishikawa2017, Gallegos2016, Polymerou2014, Paltoglou2012, Santarcangelo2015, Jurek2014, Philander2016, Walha2016, Paakkonen2016, GaoBo2016, Munezero2015, Hridoy2015, Jiang2015, Wang2015, Hagen2015, LuT2015, LuY2015, Varshney2014, Ou2014, Mostafa2013b, Mostafa2013, Lek2013, Tumasjan2010, Raja2016, Lau2014}
                                     & \cite{Singh2018, Pollacci2017, AbdullahMalak2017-1, Joyce2017, Husnain2017, Shi2017, Khuc2012, Porshnev2013, Li2017, Mukherjee2017, Chou2017, Frankenstein2016, Giachanou2016b, Feng2015, Hagen2015, DelBosque2014, Zhou2014, Souza2012, Asiaee2012}
                                     & \cite{Bansal2018, Ghiassi2018, Tasoulis2018, Wu2016c, LiW2016, Pandarachalil2015, Molina2014, Andriotis2014, Javed2014, Chen2012}
                                     & \cite{Boididou2018, Su2017, Bandhakavi2016, Saif2014c}
                                     & \cite{Goncalves2013}
                                     & \cite{Nielsen2011}
                                     & \cite{Erdmann2014}
                                     & \cite{Rathan2018, Ranjan2018, Nausheen2018, WangHongwei2018, Vo2017, Yan2017, Radhika2017, Hu2017, Kaushik2016, Akcora2010, Gupta2016, Lai2015, Dasgupta2015, Sarlan2014, Parthasarathi2012, Tian2015, Song2015, Choi2013, Park2011, Costa2014, Yanmei2015}   
                                      \\ \hline
    	\end{tabular}
	\end{center}
	\caption{Lexicon-based studies}
	\label{table:studies-lexicon}
\end{table}

%\vspace*{-7mm}

Table \ref{table:studies-lexicon} presents the number of lexicons (first row and columns titled 1 to 8) used by the lexicon-based studies (second row). The column titled ``Other/NA" refers to any other general lexicon, which does not list general lexicons mentioned in the studies such as acronym dictionaries, intensifier words\footnote{Adverbs/adverbial phrases that strengthen the meaning of other expressions and show emphasis e.g., very, extremely}, downtoner words\footnote{Words/phrases that reduce the force of another word/phrase e.g., hardly, slightly}, negation words and internet slang, and/or to studies which do not provide any information on the exact lexicons used. 
   
The majority of the lexicon-based studies used one or two lexicons, where a total of 144 state-of-the-art lexicons (55 unique ones) were used across. The following are the top six lexicons based on use:
\begin{enumerate}
	\item SentiWordNet\footnote{https://sentiwordnet.isti.cnr.it/}
	   \cite{Baccianella2010} - used in 22 studies;
	\item Hu and Liu\footnote{https://www.cs.uic.edu/\~liub/FBS/sentiment-analysis.html} 
	 \cite{Hu2004} - used in 12 studies;
	\item AFINN\footnote{https://github.com/fnielsen/afinn} \cite{Nielsen2011} and SentiStrength\footnote{http://sentistrength.wlv.ac.uk/} 
	 \cite{Thelwall2012} - used in 9 studies;
	\item MPQA - Subjectivity\footnote{\url{https://mpqa.cs.pitt.edu/lexicons/subj_lexicon/}}
	  \cite{Wilson2005} - used in 8 studies;
	\item HowNet Sentiment Analysis Word Library (HowNetSenti)\footnote{http://www.keenage.com/} - used in 6 studies;
	\item NRC Word-Emotion Association Lexicon (also known as NRC Emotion Lexicon or EmoLex)\footnote{https://www.saifmohammad.com/WebPages/NRC-Emotion-Lexicon.htm} \cite{Mohammad2010, Mohammad2013}, 
	WordNet\footnote{https://wordnet.princeton.edu/} \cite{Miller1995} and 
	Wikipedia - list of emoticons\footnote{\url{https://en.wikipedia.org/wiki/List_of_emoticons}} - used in 5 studies.
	
\end{enumerate}

In addition to the lexicons mentioned above, 19 studies used lexicons that they created as part of their work or specifically focused on creating SOM lexicons, such as \cite{Nielsen2011} who created the AFINN word list for sentiment analysis in microblogs, \cite{Javed2014} who built a bilingual sentiment lexicon for English and Roman Urdu, \cite{Santarcangelo2015} the creators of the first Italian sentiment thesaurus, \cite{Wu2016c} for Chinese sentiment analysis and \cite{Bandhakavi2016} for sentiment analysis on Twitter. 
These lexicons varied from social media focused lexicons \cite{Tian2015, Ghiassi2018, Pollacci2017}, to sentiment and/or emoticon lexicons \cite{Jurek2014, Molina2014, Khuc2012, Ranjan2018, Vo2017, Feng2015, Wang2015, Zhou2014} and extensions of existing state-of-the-art lexicons \cite{LiW2016, Pandarachalil2015, Andriotis2014}, such as \cite{LiW2016} who extended HowNetSenti with words manually collected from the internet, and \cite{Pandarachalil2015} who built a sentiment lexicon from SenticNet\footnote{https://www.sentic.net/} \cite{Cambria2020} and SentiWordNet for slang words and acronyms.
\subsubsection{Machine Learning}
\label{sssec_ml_approaches}

A total of 121 studies adopted a machine learning-based approach to perform a form of SOM, where several supervised and unsupervised algorithms were used. Table \ref{table:studies-ml} below presents the number of machine learning algorithms (first row and columns titled 1 to 7) used by the machine learning-based studies (second row). The column titled ``NA" refers to studies who do not provide any information on the exact algorithms used.  

\begin{table} [h!]
    \footnotesize
	\begin{center}
    	\begin{tabular}{ | p{1.75cm} | p{1.5cm} | p{0.75cm} | p{0.75cm} | p{0.75cm} | p{0.75cm} | p{0.75cm} | p{0.75cm} | p{0.5cm} | }
    	\hline
        \textbf{Number of machine learning algorithms}  & 1  & 2  & 3  & 4  & 5 & 6 & 7 & NA \\ \hline
        \textbf{Number of studies }       & 59 & 23 & 18 & 9  & 5 & 2 & 1 & 4  \\ \hline
        \textbf{References of studies} & \cite{Fatyanosa2018, DosSantos2018, Rout2018, LiuQuanchao2018, Katz2018, Huang2018, Halibas2018, Ignatov2017, Vo2017, Omar2017, Ducange2017, Joyce2017, Soni2017, Radhika2017, Wehrmann2017, Song2017, Li2017, Sygkounas2016, Kumar2016, Balaji2016, Severyn2016, Singh2016, Abdelrazeq2016, WangYaqi2016, Nagiwale2015, Smailovic2015, Attigeri2015, Davanzo2015, Kokkinogenis2015, Seron2015, LuT2015, Wagner2015, Liu2015, Sluban2015, LuY2015, Guerra2014, Du2014, YanG2014, Lau2014, Batista2014, Rao2014, Tapia2014, Molina2014, Abdul2014, Li2014c, Ghiassi2013, Porshnev2013, Hoang2013, Kranjc2013, Goncalves2013, Wunnasri2013, Yu2013, Weiss2013, Xiong2013, WangH2012, Xie2012, Saif2012, Bifet2011, Bollen2011}
                                     & \cite{ZhangYazhou2018, Moh2017, Nugroho2017, GuptaF2017, Shi2017, Hao2017, WangYang2016, Suresh2016, Peng2016, Ramteke2016, Shyamasundar2016, deSouzaCarvalho2016, Lu2016, Zhang2015, Esiyok2015, WangMin2014, Filice2014, Garg2014, Ou2014, Politopoulou2013, Mejova2013, Wang2013, Li2013} 
                                     & \cite{Ismail2018, Baltas2017, Yan2017, Vora2017, Sun2017, Balikas2017, Anastasia2016, LiW2016, Khalil2015, Anjaria2014, Zimmermann2014, Le2014, Neethu2013, Paltoglou2012, Zhang2011, Bifet2010, Pak2010, Go2009}
                                     & \cite{Pant2018, Wazery2018, Adibi2018, Michailidis2018, Dritsas2018, Troussas2016, Krouska2016, Rexha2016, Zhang2014}
                                     & \cite{Xiaomei2018, Sanchez2018, Pavel2017, Baecchi2016, Asiaee2012}
                                     & \cite{Celiktug2018, Raja2016}
                                     & \cite{Juneja2017}
                                     & \cite{Rezk2018, Weiss2015, Brooks2014, Sheth2014}
                                      \\ \hline
    	\end{tabular}
	\end{center}
	\caption{Machine learning-based studies}
	\label{table:studies-ml}
\end{table}

%\vspace*{-7mm}
 
In total, 239 machine learning algorithms were used (not distinct) across 117 studies (since 4 studies did not provide any information), with 235 being supervised and 4 unsupervised. It is important to note that this figure does not include any supervised/semi-supervised/unsupervised proposed algorithms by the respective authors, which algorithms shall be discussed below. 

\begin{table} [h!]
	\small
	\begin{center}
      \begin{tabular}{ | c | c | c | }
    	\hline
        \textbf{Algorithm}              & \textbf{Number of studies} & \textbf{Reference} \\ \hline 	
     	Na\"ive Bayes (NB)                  & 75       & \cite{Lewis1998}       \\ \hline 
     	Support Vector Machine (SVM)      & 71       & \cite{Cortes1995}      \\ \hline 
     	Logistic Regression (LoR)         & 16       & \cite{Mccullagh1984}   \\ \hline 
     	Decision Tree (DT)                & 15       & \cite{Quinlan1986}     \\ \hline 
     	Maximum Entropy (MaxEnt)          & 12       & \cite{Jaynes1957}      \\ \hline      		
     	Random Forest (RF)                & 9        & \cite{Breiman2001}     \\ \hline 
     	K-Nearest Neighbors (KNN)         & 7        & \cite{Altman1992}      \\ \hline      
     	SentiStrength                     & 5        & \cite{Thelwall2012}    \\ \hline
     	Conditional Random Field 	      & 4        & \cite{Lafferty2001}    \\ \hline  
     	Linear Regression (LiR)           & 4        & \cite{Cook1977}        \\ \hline    	
     	SANT optimization algorithm (SANT)& 3        & \cite{Hu2013}          \\ \hline 
     	Stochastic Gradient Descent (SGD) & 3        & \cite{Bottou2010}      \\ \hline 
     	Passive Aggressive (PA)           & 2        & \cite{Crammer2006}     \\ \hline 
     	Bootstrap Aggregating (Bagging)   & 1        & \cite{Breiman1996}     \\ \hline 
     	Bayesian Network (BN)             & 1        & \cite{Heckerman1995}   \\ \hline 
     	Conjunctive Rule Based (CRB)      & 1        & \cite{Clark1989}       \\ \hline 
     	Adaptive Boosting (AB)            & 1        & \cite{Freund1999}      \\ \hline 
     	Hidden Markov Model (HMM)         & 1        & \cite{Baum1966}        \\ \hline  	                    
     	Dictionary Learning 	          & l        & \cite{Ramirez2010}     \\ \hline 
     	SVM with NB features (NBSVM)      & 1		 & \cite{Wang2012}	      \\ \hline
     	Multiclass Classifier (MCC)       & l        & \cite{Witten2016}      \\ \hline 
        Iterative Classifier Optimizer (ICO)  & l    & \cite{Witten2016}      \\ \hline 
     	      
     \end{tabular}
	\end{center}
	\caption{Supervised machine learning algorithms}
	\label{table:studies-ml-supervised}
\end{table}

Table \ref{table:studies-ml-supervised} provides breakdown of the 235 supervised machine learning algorithms (not distinct) that were used within these studies. 
The NB and SVM algorithms are clearly the most popular in this domain, especially for text classification. With respect to the former, it is important to note that 20 out of the 75 studies used the Multinomial NB (MNB), which model is usually utilised for discrete counts i.e., the number of times a given term (word or token) appears in a document. The other 55 studies made use of the Multi-variate Bernoulli NB (MBNB) model, which is based on binary data, where every token in a feature vector of a document is classified with the value of 0 or 1. As for SVM, this method looks at the given data and sorts it in two categories (binary classification). If multi-class classification is required, the Support Vector Classification (SVC)\footnote{https://scikit-learn.org/stable/modules/generated/sklearn.svm.SVC.html\#sklearn.svm.SVC}, NuSVC\footnote{https://scikit-learn.org/stable/modules/generated/sklearn.svm.NuSVC.html\#sklearn.svm.NuSVC} or LinearSVC\footnote{https://scikit-learn.org/stable/modules/generated/sklearn.svm.LinearSVC.html\#sklearn.svm.LinearSVC} algorithms are usually applied, where the ``one-against-one" approach is implemented for SVC and NuSVC, whereas the ``one-vs-the-rest" multi-class strategy is implemented for LinearSVC.

The LoR statistical technique is also widely used in machine learning for binary classification problems. In total, 16 studies from the ones analysed, made use of this algorithm. DT learning has also been very much in use, which model uses a DT for both classification and regression problems. There are various algorithms in how a DT is built, with 2 studies using the C4.5 \cite{Quinlan1993}  
-- an extension of Quinlan's Iterative Dichotomiser 3 (ID3) algorithm, used for classification purposes, 3 studies using J48, a simple C4.5 DT for classification (Weka's implementation\footnote{http://weka.sourceforge.net/doc.dev/weka/classifiers/trees/J48.html}), 2 using the Hoeffding Tree \cite{Hulten2001} and the other 8 using the basic ID3 algorithm. 

MaxEnt, used by 12 studies, is a probabilistic classifier that is also used for text classification problems, such as sentiment analysis. More specifically, it is generalisation of LoR for multi-class scenarios \cite{Yu2011}. RF was used in 9 studies, where this supervised learning algorithm --which can be used for both classification and regression tasks-- creates a forest (which is an ensemble of DTs) and makes it somehow random. Moreover, 7 studies used the KNN algorithm, one of the simplest classification algorithms where no learning is required, since the model structure is determined from the entire dataset. 

The SentiStrength algorithm, utilised by 5 studies \cite{Goncalves2013, LuY2015, Baecchi2016, Yan2017, ZhangYazhou2018}, can be used in both supervised and unsupervised cases, since the authors developed a version for each learning case. Conditional Random Fields, used by 4 studies \cite{Pak2010, Zhang2014, WangYang2016, Hao2017}, are a type of discriminative classifier that model the decision boundary amongst different classes, whereas LiR was also used by 4 studies \cite{Bollen2011, Pavel2017, Adibi2018, Xiaomei2018}. Moreover, 3 studies each used the SANT \cite{Ou2014, LuT2015, Xiaomei2018} and SGD \cite{Bifet2010, Juneja2017, Sanchez2018} algorithms, with the former being mostly used for comparison purposes to the proposed approaches by the respective authors. 

In addition, the PA algorithm was used in 2 studies \cite{Li2014c, Filice2014}. In the case of the former \cite{Li2014c}, this algorithm was used in a collaborative online learning framework to automatically classify whether a post is emotional or not, thereby overcoming challenges faced by the diversity of microblogging styles which increase the difficulty of classification. The authors in the latter study \cite{Filice2014} extend the budgeted PA algorithm to enable robust and efficient natural language learning processes based on semantic kernels. The proposed online learning learner was applied to two real world linguistic tasks, one of which was sentiment analysis.

Nine other algorithms were used by 7 different studies, namely: Bagging \cite{Sygkounas2016}, BN \cite{Lu2016}, CRB \cite{Raja2016}, AB \cite{Raja2016}, HMM \cite{Zhang2014}, Dictionary Learning \cite{Asiaee2012}, NBSVM \cite{Sun2017}, MCC \cite{Celiktug2018} and ICO \cite{Celiktug2018}. 

In terms of unsupervised machine learning algorithms, 4 were used in 2 of the 80 studies that used a machine learning-based technique. Suresh and Raj S. used the K-Means (KM) \cite{Lloyd1982} and Expectation Maximization \cite{Dempster1977} clustering algorithms in \cite{Suresh2016}. Both were used for comparison purposes to an unsupervised modified fuzzy clustering algorithm proposed by authors. The proposed algorithm produced accurate results without manual processing, linguistic knowledge or training time, which concepts are required for supervised approaches. Baecchi et al. \cite{Baecchi2016} used two unsupervised algorithms, namely Continuous Bag-Of-Word (CBOW) \cite{Mikolov2013} and Denoising Autoencoder (DA) \cite{Vincent2008} (the SGD and backpropagation algorithms were used for the DA learning process), and supervised ones, namely LoR, SVM and SentiStrength, for constructing their method and comparison purposes. They considered both textual and visual information in their work on sentiment analysis of social network multimedia. Their proposed unified model (CBOW-DA-LoR) works in both an unsupervised and semi-supervised manner, whereby learning text and image representation and also the sentiment polarity classifier for tweets containing images.
 
Other studies proposed their own algorithms, with some of the already established algorithms discussed above playing an important role in their implementation and/or comparison. Zimmermann et al. proposed a semi-supervised algorithm, the S*3Learner \cite{Zimmermann2014} which suits changing opinion stream classification environments, where the vector of words evolves over time, with new words appearing and old words disappearing. In \cite{Severyn2016}, Severyn et al. defined a novel and efficient tree kernel function, the Shallow syntactic Tree Kernel, for multi-class supervised sentiment classification of online comments. This study focused on YouTube which is multilingual, multimodal, multidomain and multicultural, with the aim to find whether the polarity of a comment is directed towards the source video, product described in the video or another product. Furthermore, Ignatov and Ignatov \cite{Ignatov2017} presented a novel DT-based algorithm, a Decision Stream, where Twitter sentiment analysis was one of several common machine learning problems that it was evaluated on. Lastly, the authors in \cite{Fatyanosa2018}, enhanced the ability of the NB classifier with an optimisation algorithm, the Variable Length Chromosome Genetic Algorithm (VLCGA), thereby proposing VLCGA-NB for Twitter sentiment analysis.

Moreover, the following 13 studies proposed an ensemble method or evaluated ensemble-based classifiers: 
\begin{itemize}
	\item \cite{Celiktug2018} used two ensemble learning methods in RF and MCC (amongst other machine learning algorithms), for sentiment classification of Twitter datasets;
	\item \cite{Yan2017} presented two ensemble learners built on four off-the-shelf classifiers, for Twitter sentiment classification;
	\item \cite{ZhangYazhou2018, Adibi2018, Celiktug2018, Vora2017, Lu2016, Rexha2016, Xie2012, Zhang2011} used the RF ensemble learning method in their work;
    \item \cite{Troussas2016} evaluated the most common ensemble methods that can be used for sentiment analysis on Twitter datasets;
	\item \cite{Sygkounas2016} proposed an ensemble system composed on five state-of-the-art sentiment classifiers;
	\item \cite{Le2014} used multiple oblique decision stumps classifiers to form an ensemble of classifiers, which is more accurate than a single one for classifying tweets;
	\item \cite{Neethu2013} used an ensemble classifier (and single algorithm classifiers) for sentiment classification.
\end{itemize} 
Ensembles created usually result in providing more accurate classification answers when compared to individual classifiers, i.e., classic learning approaches. In addition, ensembles reduce the overall risk of choosing a wrong classifier especially when applying it on a new dataset \cite{DaSilva2014}.

\subsubsection{Deep Learning}
\label{sssec_dl_approaches}

Deep learning is a subset of machine learning based on Artificial Neural Networks (ANNs) --algorithms inspired by the human brain-- where there are connections, layers and neurons for data to propagate. A total of 35 studies adopted a deep learning-based approach to perform a form of SOM, where supervised and unsupervised algorithms were used. Twenty six (26) of the studies made use of 1 deep learning algorithm, with 5 utilising 2 algorithms and 2 studies each using 3 and 4 algorithms, respectively. Table \ref{table:studies-dl} provides breakdown of the 50 deep learning algorithms (not distinct) used within these studies. 

\begin{table} [h!]
	\small
	\begin{center}
      %\begin{tabular}{ | c | c | c | }
      \begin{tabular}{ | p{7.25cm} | p{2cm} | p{1.5cm} | }
    	\hline
        \textbf{Algorithm}                     & \textbf{Number of studies} & \textbf{Reference} \\ \hline 	
     	
     	Long Short-Term Memory (LSTM)          & 13       &	\cite{Hinton2012}      \\ \hline 
	    Convolutional Neural Network (CNN)     & 12		  & \cite{Lecun1990}	   \\ \hline 		
		Recurrent Neural Network (RNN)		   & 8    	  & \cite{Graves2005}	   \\ \hline 
     	ANN								       & 5        & \cite{Mcculloch1943}   \\ \hline 
     	Recursive Neural Tensor Network (RNTN) & 3        &	\cite{Socher2013}      \\ \hline 
     	Bidirectional Long Short-Term Memory (BLSTM) & 3  & \cite{Graves2005}      \\ \hline
     	Multilayer Perceptron (MLP) 		   & 2		  & \cite{Hornik1989}	   \\ \hline
     	Autoencoder (AE)					   & 2  	  & \cite{Rumelhart1985}   \\ \hline
     	Gated Recurrent Units (GRU)  		   & 1		  & \cite{Greff2017}	   \\ \hline
     	Dynamic Architecture for ANN (DAN2)    & 1	      & \cite{Ghiassi2005}     \\ \hline
     \end{tabular}
	\end{center}
	\caption{Deep learning algorithms}
	\label{table:studies-dl}
\end{table}

LSTM, a prominent variation of the RNN which makes it easier to remember past data in memory, was used in 13 studies \cite{Yan2018, SunXiaoZhang2018, Sanyal2018, Ameur2018, Wazery2018, LiD2018, ChenN2018, ChenYuxiao2018, Sun2017, Hu2017, Shi2017, WangYang2016, Yan2016}, thus making it the most popular deep learning algorithm amongst the evaluated studies. Three further studies \cite{Ameur2018, Balikas2017, WangYang2016} used the BLSTM, an extension of the traditional LSTM which can improve model performance on sequence classification problems. In particular, a BLSTM was used in \cite{Balikas2017} to improve the performance of fine-grained sentiment classification, which approach can benefit sentiment expressed in different textual types (e.g., tweets and paragraphs), in different languages and different granularity levels (e.g., binary and ternary). Similarly, Wang et al. \cite{WangYang2016} proposed a language-independent method based on BLSTM models for incorporating preceding microblogs for context-aware Chinese sentiment classification.

The CNN algorithm --a variant of ANN-- is made up of neurons that have learnable weights and biases, where each neuron receives an input, performs a dot product and optionally follows it with non-linearity. %CNNs are made up of two parts: feature learning and classification. 
In total, 12 studies \cite{SunXiaoZhang2018, Ochoa2018, Ameur2018, Adibi2018, ChenN2018, Napitu2017, Shi2017, Wehrmann2017, Zhang2017, Stojanovski2015, WangYaqi2016, Severyn2015} made use of this algorithm. Notably, the authors in \cite{Wehrmann2017} propose a language-agnostic translation-free method for Twitter sentiment analysis.

RNNs, a powerful set of ANNs useful for processing and recognising patterns in sequential data such as natural language, were used in 8 studies \cite{Yan2018, Ochoa2018, Pineiro2018, Wazery2018, Pavel2017, Shi2017, Yan2016, WangYang2016}. One study in particular \cite{Averchenkov2015}, considered a novel approach to aspect-based sentiment analysis of Russian social networks based on RNNs, where the best results were obtained by using a special network modification, the RNTN. Two further studies \cite{LuY2015, Sygkounas2016} also used this algorithm (RNTN) in their work.

Five other studies \cite{Arslan2018, Anjaria2014, Du2014, Politopoulou2013, Zhang2011} used a simple type of ANN, such as the feedforward neural network. Moreover, the MLP, a class of feedforward ANN, was used in 2 studies \cite{Chen2018, Ramadhani2017}. Similarly, 2 studies \cite{Yan2018, Ameur2018} proposed methods based on the AE unsupervised learning algorithm which is used for representation learning. Lastly, one study each made use of the GRU \cite{WangYang2016} and DAN2 \cite{Ghiassi2013} algorithms.

Some studies used several types of ANNs in their work. The authors in \cite{Ameur2018} used multiple methods based on AE, CNN, LSTM and BLSTM for sentiment polarity classification and \cite{WangYang2016} used RNN, LSTM, BLSTM and GRUs models. In \cite{Yan2018}, Yan et al. used learning methods based on RNN, LSTM and AE for comparison with the proposed learning framework for short text classification, and \cite{Shi2017} proposed an improved LSTM which considers user-based and content-based features and used CNN, LSTM and RNN models for comparison purposes. Furthermore, \cite{Ochoa2018} made use of CNN and RNN deep learning algorithms for tweet sentiment analysis, \cite{Wazery2018, Yan2016} used the RNN and LSTM, whereas \cite{SunXiaoZhang2018, ChenN2018} proposed new models based on CNN and LSTM. 

\subsubsection{Statistical}
\label{sssec_stat_approaches}

A total of 9 studies \cite{WangYili2018, Kitaoka2017, Arslan2017, Raja2016, YangC2014, Bukhari2016, Zhang2015, Petz2013, Supriya2016} adopted a statistical approach to perform a form of SOM. In particular, one of the approaches proposed in \cite{Arslan2017} uses the term frequency-inverse document frequency (tf-idf) \cite{Salton1986} numerical statistic to find out the important words within a tweet, to dynamically enrich Twitter specific dictionaries created by the authors. The tf-idf is also one of several statistical-based techniques used in \cite{WangYili2018} for comparing the proposed novel feature weighting approach for Twitter sentiment analysis. Moreover, \cite{Raja2016} focuses on a statistical sentiment score calculation technique based on adjectives, whereas the authors in \cite{YangC2014} use a variation of the point-wise mutual information to measure the opinion polarity of an entity and its competitors, which method is different from the traditional opinion mining way.

\subsubsection{Probabilistic}
\label{sssec_prob_approaches}

A total of 6 studies \cite{Bhattacharya2017, Baecchi2016, Ou2014, Ragavi2014, YanG2014, Lek2013} adopted a probabilistic approach to perform a form of SOM. In particular, \cite{Ou2014} propose a novel probabilistic model in the Content and Link Unsupervised Sentiment Model, where the focus is on microblog sentiment classification incorporating link information, namely behaviour, same user and friend.

\subsubsection{Fuzziness}
\label{sssec_fuzzy_approaches}

Two studies \cite{DAsaro2017, DelBosque2014} adopted a fuzzy-based approach to perform a form of SOM. D'Asaro et al. \cite{DAsaro2017} present a sentiment evaluation and analysis system based on fuzzy linguistic textual analysis. In \cite{DelBosque2014}, the authors assume that aggressive text detection is a sub-task of sentiment analysis, which is closely related to document polarity detection given that aggressive text can be seen as intrinsically negative. This approach considers the document's length and the number of swear words as inputs, with the output being an aggressiveness value between 0 and 1.

\subsubsection{Rule-based}
\label{sssec_rule_approaches}

In total, 4 studies \cite{ElHaddaoui2018, Zhang2014, Min2013, Bosco2013} adopted a rule-based approach to perform a form of SOM. Notably, Bosco et al. \cite{Bosco2013} applied an approach for automatic emotion annotation of ironic tweets. This relies on sentiment lexicons (words and expressions) and sentiment grammar expressed by compositional rules. 

\subsubsection{Graph}
\label{sssec_graph_approaches}

Four studies \cite{Dritsas2018, Vilarinho2018, ChenF2015, Rabelo2012} adopted a graph-based approach to perform a form of SOM. The study in \cite{Vilarinho2018} presents a word graph-based method for Twitter sentiment analysis using global centrality metrics over graphs to evaluate sentiment polarity. In \cite{Dritsas2018}, a graph-based method is proposed for sentiment classification at a hashtag level. Moreover, the authors in \cite{ChenF2015} compare their proposed multimodal hypergraph-based microblog sentiment prediction approach with a combined hypergraph-based method \cite{Huang2010}. Lastly, \cite{Rabelo2012} used link mining techniques to infer the opinions of users.  

\subsubsection{Ontology}
\label{sssec_ontology_approaches}

Two studies \cite{Lau2014, Kontopoulos2013} adopted an ontology-based approach to perform a form of SOM. In particular, the technique developed in \cite{Kontopoulos2013} performs more fine-grained sentiment analysis of tweets where each subject within the tweets is broken down into a set of aspects, with each one being assigned a sentiment score.

\subsubsection{Hybrid}
\label{sssec_hybrid_approaches}

Hybrid approaches are very much in demand for performing different opinion mining tasks, where 244 unique studies (out of 465) adopted this approach and produced a total of 282 different techniques\footnote{Several studies contain multiple hybrid methods, which are included in the respective hybrid approach total}

Tables \ref{table:studies-hybrid-2} and \ref{table:studies-hybrid-3-4} lists these studies, together with the type of techniques used for each. In total, there were 38 different hybrid approaches across the analysed studies.

%\vspace*{-4mm}

\begin{table} [htp]
    \footnotesize
	\begin{center}
    	\begin{tabular}{ | p{0.25cm} | p{0.40cm} | p{0.40cm} | p{0.25cm} | p{0.25cm} | p{0.25cm} | p{0.25cm} | p{0.25cm} | p{0.30cm} | p{0.70cm} | p{5.35cm} |}
    	\hline
     	             \textbf{Lx} & \textbf{ML} & \textbf{DL} & \textbf{St} & \textbf{Pr} & \textbf{Fz} & \textbf{Rl} & \textbf{Gr} & \textbf{On} & \textbf{Total} & \textbf{Studies} \\ \hline
     	             \cmark & \cmark & & & & & & & & 114 & \cite{ZhangYazhou2018, Yan2018, Pollacci2017, Jin2017, Hong2018, Calvo2018, Rathan2018, Saleena2018, Yan2017, Katiyar2018, Gandhe2018, Al2018, PaiPing2018, Goel2018, Sahni2017, Ahuja2017, Fatyanosa2017, Singh2018, AbdullahNSD2017, Lee2017, Bouchlaghem2016, Sharma2016, Sankaranarayanan2016, Kanavos2016, Gallegos2016, Koto2015-1, Buscaldi2015, Tsytsarau2014, Yuan2014, Bravo2013, ZhangL2013, Xu2012, Jianqiang2017, Qaisi2016, Zimbra2016, Jianqiang2016, You2016, Bravo2016, Zhao2016, LiM2016, Deshwal2016, Jianqiang2015a, ChenX2015, LiW2016, Fersini2015, Abdelwahab2015, YangA2015, YangY2015, ChenF2015, Kanakaraj2015, Jianqiang2015b, Koto2015-2, Wu2015, Shukri2015, Sahu2015, Lewenberg2015, Cho2014, Sui2012, Karyotis2017, Lim2017, Pandey2017, Burnap2016, Lima2015, Poria2014, Bravo2014, DaSilva2014, Gambino2016, Yan2016, Jiang2015, Nguyen2016, Aboluwarin2016, Zainuddin2016b, Flaes2016, Koto2015-3, Hagen2015, Castellucci2015a, Sanborn2015, Castellucci2015b, ChenC2015, Mansour2015, DelBosque2014, Han2015, Yuan2015, Buddhitha2015, Ji2015, Zhou2014, WangMingqiang2014, Tsakalidis2014, Porshnev2014a, SuZ2014, YanB2014, Goncalves2013, Porshnev2014b, Sun2014, Pla2014, WangDongfang2014, Bao2014, Zhu2013, Jiang2013, Cui2013, Khuc2012, Bermingham2010, WangW2012, Montejo2014, Ortigosa2014, Rui2013, Reyes2013, Kouloumpis2011, Bakliwal2013, Vu2012, Agarwal2011, Hernandez2014, Thelwall2011, Thelwall2010}\\ \hline 
     	             \cmark & & \cmark & & & & & & & 12 & \cite{Jin2017, Baccouche2018, Karyotis2017, Zimbra2016, Er2016, Koto2015-3, Bravo2014, DelBosque2014, Porshnev2014a, Porshnev2014b, Tang2013, Thelwall2010} \\ \hline  
     	             \cmark & & & \cmark & & & & & & 22 & \cite{ZhangYazhou2018, Wan2018, Sangameswar2017, Rout2018, Rout2017, Bansal2018, Satapathy2017, Tago2018, Fatyanosa2017, Sachdeva2018, Zhou2017, Le2017, Azzouza2017, Giachanou2016c, GaoFei2016, Lu2016, Orellana2015, Tan2014, Khan2014, Orellana2013, Blenn2012, Zhang2012} \\ \hline  
     	             \cmark & & & & \cmark & & & & & 3 & \cite{Huang2015, Yang2013, Lek2013} \\ \hline  
     	             \cmark & & & & & \cmark & & & & 4 & \cite{Ismail2018, Cotfas2017, Kao2018, Dragoni2018} \\ \hline
     	             \cmark & & & & & & \cmark & & & 9 & \cite{Dambhare2017, ChenYuxiao2018, Kamyab2018, Mishra2018, Gambino2016, WangZhitao2014, Saif2014a, Wang2013, Maynard2011} \\ \hline  
     	             \cmark & & & & & & & \cmark & & 4 & \cite{ChenF2015, Rill2014, Bliss2012, Cui2011} \\ \hline  
     	             \cmark & & & & & & & & \cmark & 2 & \cite{Cotfas2015, Delcea2014} \\ \hline  
     	             & \cmark & \cmark & & & & & & & 7 & \cite{Yan2018, Stojanovski2018, Ameur2018, Sun2017, Prusa2015, Yanmei2015, Cai2015}  \\ \hline  
     	             & \cmark & & \cmark & & & & & & 21 & \cite{WangYili2018, Saidani2017, Sabuj2017, Ismail2018, Hanafy2018, Elouardighi2017, Effrosynidis2017, Ameur2018, Symeonidis2018, Rezaei2017, Elzayady2018, Rinaldi2017, Coyne2017, Setiawan2018, Dedhia2017, Alzahrani2018, Elbagir2018, Mishra2018, Ramadhani2016, Trung2013, Taddy2013} \\ \hline  
     	             & \cmark & & & \cmark & & & & & 3 & \cite{Vo2017, Sihwi2018, Caschera2016} \\ \hline   
     	             & \cmark & & & & & \cmark & & & 2 & \cite{Mumu2014, ZhangH2013} \\ \hline  
     	             & \cmark & & & & & & \cmark & & 3 & \cite{LuT2015, Ou2014, Lek2013} \\ \hline 
     	             & & \cmark & \cmark & & & & & & 3 & \cite{Hanafy2018, Symeonidis2018, Coyne2017} \\ \hline  
     	             & & \cmark & & \cmark & & & & & 1 & \cite{Haldenwang2018} \\ \hline              
     	            % & & & \cmark & & & & \cmark & & &  \\ \hline  
     	             & & & & \cmark & \cmark & & & & 1 & \cite{Mukkamala2014-2} \\ \hline
     	             & & & & & \cmark & \cmark & & & 1 & \cite{Karyotis2017}  \\ \hline
     	             & & & & & \cmark & & \cmark & & 1 & \cite{Mukkamala2014-1}  \\ \hline           
    	\end{tabular}
	\end{center}
	\caption{Studies adopting a hybrid approach consisting of two techniques}
	\label{table:studies-hybrid-2}
\end{table}

%\vspace*{-8mm}

%\vspace*{-4mm}

\begin{table} [htp]
    \small
	\begin{center}
    	\begin{tabular}{ | p{0.30cm} | p{0.45cm} | p{0.40cm} | p{0.30cm} | p{0.30cm} | p{0.30cm} | p{0.30cm} | p{0.30cm} | p{0.40cm} | p{0.80cm} | p{4.80cm} |}
    	\hline
     	             \textbf{Lx} & \textbf{ML} & \textbf{DL} & \textbf{St} & \textbf{Pr} & \textbf{Fz} & \textbf{Rl} & \textbf{Gr} & \textbf{On} & \textbf{Total} & \textbf{Studies} \\ \hline
     	             \cmark & \cmark & \cmark & & & & & & & 3 & \cite{Cao2018, Hassan2013, Kalayeh2015} \\ \hline
     	             \cmark & \cmark & & \cmark & & & & & & 21 & \cite{Vo2017, Villegas2018, Konate2018, Giachanou2017, Alharbi2017, Saleena2018, Ghiassi2018, Tellez2017, Simoes2017, Lavanya2017, LiYujiao2018, Permatasari2018, Fitri2018, Bouazizi2018, Rai2018, Wijayanti2017, Xia2017, Jianqiang2018, Bouazizi2017, ChenP2015, Pei2014}  \\ \hline
     	             \cmark & \cmark & & & \cmark & & & & & 2 & \cite{Ortis2018, Lek2013} \\ \hline
     	             \cmark & \cmark & & & & & \cmark & & & 12 & \cite{Moh2017, Siddiqua2016, Liu2016, Zainuddin2016a, Poria2016, Souza2016, Chikersal2015, Shi2013, Maeda2012, Li2014, Prabowo2009, Thelwall2012}  \\ \hline
     	              \cmark & & \cmark & \cmark & & & & & & 7 & \cite{Villegas2018, LiuQuanchao2018, Konate2018, Ghosal2018, Ghiassi2018, LiYujiao2018, Jianqiang2018}  \\ \hline
     	              %\cmark & & \cmark & & \cmark & & & & & 1 & \cite{}  \\ \hline
     	             \cmark & & \cmark & & & & \cmark & & & 2 & \cite{Saini2018, Poria2016}  \\ \hline
     	             \cmark & \cmark & & & & & & \cmark & & 2 & \cite{Weichselbraun2017, Jiang2011}  \\ \hline
     	             \cmark & \cmark & & & & & & & \cmark & 1 & \cite{Ji2016} \\ \hline
     	             \cmark & & & \cmark & \cmark & & & & & 1 & \cite{ZhangYazhou2018} \\ \hline   
     	             \cmark & & & \cmark & & \cmark & & & & 1 & \cite{Dragoni2018} \\ \hline   	             	               	             
     	             \cmark & & & \cmark & & & \cmark & & & 3 & \cite{Asghar2018, Gao2015, Unankard2014} \\ \hline
     	             & \cmark & \cmark & \cmark & & & & & & 2 & \cite{Hanafy2018, Ameur2018} \\ \hline	             
     	             & \cmark & & \cmark & \cmark & & & & & 1 & \cite{Tong2017} \\ \hline
     	             & \cmark & & \cmark & & & \cmark & & & 1 & \cite{Samoylov2014} \\ \hline 
     	             & \cmark & & \cmark & & & & \cmark & & 2 & \cite{Xiaomei2018, Tan2011}  \\ \hline 
     	             & \cmark & & & & \cmark & \cmark & & & 1 & \cite{Nivetha2016} \\ \hline      	   
     	              \cmark & \cmark & \cmark & \cmark & & & & & & 2 & \cite{Chen2017, Alharbi2017} \\ \hline   
     	          \cmark & \cmark & & \cmark & \cmark & & & & & 1 & \cite{Vo2017} \\ \hline      	                 
     	             \cmark & \cmark & & \cmark & & & \cmark & & & 3 & \cite{Zainuddin2018, Saif2014b, Korenek2014} \\ \hline
     	             \cmark & \cmark & & & & & \cmark & \cmark & & 1 & \cite{Kuo2016} \\ \hline   
    	\end{tabular}
	\end{center}
	\caption{Studies adopting a hybrid approach consisting of three and four techniques}
	\label{table:studies-hybrid-3-4}
\end{table}

%\vspace*{-8mm}

The majority of these studies used two different techniques (213 out of 282) --see Table \ref{table:studies-hybrid-2}-- within their hybrid approach, whereas 62 used three and 7 studies used four different techniques --see Table \ref{table:studies-hybrid-3-4}. 

The Lexicon and Machine Learning-based techniques were mostly used, where they accounted for 40\% of the hybrid approaches, followed by Lexicon and Statistical-based (7.8\%), Machine Learning and Statistical-based (7.4\%), and Lexicon, Machine Learning and Statistical-based (7.4\%) techniques.

Moreover, out of the 282 hybrid approaches, 232 used lexicons, 205 used Machine Learning and 39 used Deep Learning. These numbers reflect the importance of these three techniques within the SOM research and development domain. In light of these, a list of lexicons, machine learning and deep learning algorithms used in these studies have been compiled, similar to Sections \ref{sssec_lexicon_approaches}, \ref{sssec_ml_approaches} and \ref{sssec_dl_approaches} above. The lexicons, machine learning and deep learning algorithms quoted below were either used in the proposed method/s and/or for comparison purposes in the respective studies. 

In terms of state-of-the-art lexicons, these total 403 within the studies adopting a hybrid approach. The top ones align with the results obtained from the lexicon-based approaches in Section \ref{sssec_lexicon_approaches} above. The following are the lexicons used for more than ten times across the hybrid approaches:

\begin{enumerate}
	\item SentiWordNet - used in 51 studies;
	\item MPQA - Subjectivity - used in 28 studies;
	\item Hu and Liu - used in 25 studies;
	\item WordNet - used in 24 studies;
	\item AFINN - used in 22 studies;
	\item SentiStrength - used in 21 studies;		
	\item HowNetSenti - used in 15 studies;
	\item NRC Word-Emotion Association Lexicon - used in 13 studies;
	\item NRC Hashtag Sentiment Lexicon\footnote{http://saifmohammad.com/WebPages/lexicons.html} \cite{MohammadKZ2013} - used in 12 studies; 
	\item SenticNet, 
	Sentiment140 Lexicon (also known as NRC Emoticon Lexicon)\footnote{http://saifmohammad.com/WebPages/lexicons.html} \cite{MohammadKZ2013}, National Taiwan University Sentiment Dictionary (NTUSD) \cite{Ku2006} and Wikipedia list of emoticons - used 11 studies.
\end{enumerate}

Further to the quoted lexicons, 49 studies used lexicons that they created as part of their work. Some studies composed their lexicons from emoticons/emojis that were extracted from a dataset \cite{Cao2018, LiYujiao2018, Azzouza2017, Zimbra2016, You2016, ChenF2015, Porshnev2014b, Cui2011, Zhang2012, Vu2012}, combined publicly available emoticon lexicons/lists \cite{Siddiqua2016} or mapped emoticons to their corresponding polarity \cite{Tellez2017}, and others \cite{GaoFei2016, Souza2016, SuZ2014, YanB2014, Tang2013, Cui2011, Zhang2012, Li2014} used seed/feeling/emotional words to establish a microblog typical emotional dictionary. Additionally, some authors constructed or used sentiment lexicons \cite{ZhangYazhou2018, Vo2017, Rout2017, Jin2017, Ismail2018, Yan2017, Katiyar2018, Al2018, AbdullahNSD2017, Liu2016, Sahu2015, Cho2014, WangZhitao2014, ChenP2015, Jiang2013, Cui2013, Khuc2012, Montejo2014, Rui2013} some of which are domain or language specific \cite{Konate2018, Hong2018, Chen2017, Zhao2016, Lu2016, Zhou2014, Porshnev2014a}, others that extend state-of-the-art lexicons \cite{LiM2016, LiW2016, Koto2015-1}, and some who made them available to the research community \cite{Cotfas2017, Castellucci2015b} such as the Distributional Polarity Lexicon\footnote{http://sag.art.uniroma2.it/demo-software/distributional-polarity-lexicon/}. 

%\vspace*{-5mm}

\begin{table} [htp] 
   \small
	\begin{center}
      \begin{tabular}{ | p{6.25cm} | p{2.5cm} | p{2cm} | }
    	\hline
        \textbf{Algorithm} & \textbf{Learning type} & \textbf{Studies \#}  \\ \hline    	
     	SVM                & Sup                  & 130  \\ \hline 
     	NB                 & Sup                  & 96  \\ \hline     	
     	LoR                & Sup                  & 34  \\ \hline 
     	DT                 & Sup                  & 27  \\ \hline 
     	RF                 & Sup                  & 21  \\ \hline 
     	MaxEnt             & Sup                  & 15   \\ \hline      	
     	SentiStrength      & Sup / Unsup          & 13   \\ \hline
     	LiR                & Sup                  & 8   \\ \hline     
     	KNN                & Sup                  & 5   \\ \hline   
     	AB                 & Sup                  & 5   \\ \hline    	 	       
     	BN                 & Sup                  & 3   \\ \hline      	
     	Support Vector Regression (SVR) & Sup     & 3   \\ \hline 
     	SANT               & Sup                  & 3   \\ \hline  
     	KM                 & Unsup                & 2   \\ \hline  
     	Repeated Incremental Pruning to Produce Error Reduction (RIPPER) & Sup                  & 2   \\ \hline     
     	HMM                & Sup                  & 1   \\ \hline	
     	Extremely Randomised Trees & Sup    & 1   \\ \hline           	
     	Least Median of Squares Regression  & Sup    & 1   \\ \hline      	
     	Maximum Likelihood Estimation & Sup & 1   \\ \hline 
     	Hyperpipes         & Sup                  & 1   \\ \hline
     	Extreme Learning Machine & Sup      & 1   \\ \hline  
     	Domain Adaptation Machine & Sup     & 1   \\ \hline       		
     	Affinity Propagation & Unsup         & 1   \\ \hline  
     	Multinomial inverse regression & Unsup & 1   \\ \hline  
     	Apriori            & Sup / Unsup      	  & 1   \\ \hline 
     	Distant Supervision & Semi-sup       & 1   \\ \hline  
     	Label Propagation  & Semi-sup         & 1   \\ \hline  
     	SGD  			   & Sup				  & 1 	\\ \hline 
     	NBSVM			   & Sup				  & 1 	\\ \hline
     \end{tabular}
	\end{center}
	\caption{Machine learning algorithms used in the studies adopting a hybrid approach}
	\label{table:studies-hybrid-ml}
\end{table}

%\vspace*{-7mm}

Table \ref{table:studies-hybrid-ml} below presents a list of machine learning algorithms --in total 381 in 197 studies-- that were used within the hybrid approaches. The first column indicates the algorithm, the second lists the type of learning algorithm, in terms of Supervised (Sup), Unsupervised (Unsup) and Semi-supervised (Semi-sup), and the last column lists the total number of studies using each respective algorithm. 
%TO CONT
The SVM and NB algorithms were mostly used in supervised learning, which result corresponds to the machine learning-based approaches in Section \ref{sssec_ml_approaches} above. With respect to the latter, 76 studies used the MBNB algorithm, 19 studies the MNB and 1 study the Discriminative MNB. Moreover, the LoR, DT --namely the basic ID3 (10 studies), J48 (5 studies), C4.5 (5 studies), Classification And Regression Tree (3 studies), Reduced Error Pruning (1 study), DT with AB (1 study), McDiarmid Tree \cite{Mcdiarmid1989} (1 study) and Hoeffding Tree (1 study) algorithms, RF, MaxEnt and SentiStrength (used in both supervised and unsupervised settings) algorithms were also in various studies. 
Notably, some additional algorithms from the ones used in the machine learning-based approaches in Section \ref{sssec_ml_approaches} above, were used in a hybrid approach, in particular, SVR \cite{Drucker1997}, Extremely Randomised Trees \cite{Geurts2006},  Least Median of Squares Regression \cite{Rousseeuw1984}, Maximum Likelihood Estimation \cite{Fisher1925}, Hyperpipes \cite{Witten2016}, Extreme Learning Machine \cite{Huang2006}, Domain Adaptation Machine \cite{Duan2009}, RIPPER \cite{Cohen1995}, Affinity Propagation \cite{Frey2007}, Multinomial inverse regression \cite{Taddy2013}, Apriori \cite{Agrawal1994}, Distant Supervision \cite{Go2009} and Label Propagation \cite{Zhu2002}. 

Given that deep learning is a subset of machine learning, the algorithms used within the hybrid approaches are presented below. In total, 36 studies used the following deep learning algorithms: 
\begin{itemize}
\item CNN - used in 16 studies \cite{Yan2018, Stojanovski2018, Konate2018, Hanafy2018, Haldenwang2018, Ghosal2018, Chen2017, Ameur2018, Alharbi2017, Symeonidis2018, Saini2018, Jianqiang2018, Baccouche2018, Cai2015, Kalayeh2015, Yanmei2015};
\item ANN - used in 8 studies \cite{LiYujiao2018, Karyotis2017, Poria2016, Er2016, Koto2015-3, Porshnev2014a, Porshnev2014b, Hassan2013};
\item LSTM - used in 7 studies \cite{Yan2018, Konate2018, Hanafy2018, Ghosal2018, Ameur2018, Sun2017, Baccouche2018};
\item MLP - used in 7 studies \cite{Villegas2018, Ghosal2018, Coyne2017, Karyotis2017, Bravo2014, DelBosque2014, Thelwall2010};
\item RNN - used in 4 studies \cite{Yan2018, LiuQuanchao2018, Baccouche2018, Yanmei2015};
\item AE - used in 2 studies \cite{Yan2018, Ameur2018};
\item BLSTM - used in 2 studies \cite{Konate2018, Ameur2018};
\item DAN2 - used in 2 studies \cite{Ghiassi2018, Zimbra2016};
\item Deep Belief Network \cite{Hinton2006}, a probabilistic generative model that is composed of multiple layers of stochastic, latent variables - used in 2 studies \cite{Jin2017, Tang2013};
\item GRU - used in 1 study \cite{Cao2018};
\item Generative Adversarial Networks (GAN) \cite{Goodfellow2014}, are deep neural net architectures composed of a two networks, a generator and a discriminator, pitting one against the other - used in 1 study \cite{Cao2018};
\item Conditional GAN \cite{Mirza2014}, a conditional version of GAN that can be constructed by feeding the data that needs to be conditioned on both the generator and discriminator - used in 1 study \cite{Cao2018};
\item Hierarchical Attention Network, a neural architecture for document classification \cite{Yang2016}, used in 1 study \cite{LiuQuanchao2018}.
\end{itemize}

Further to the quoted algorithms, 22 studies \cite{Hong2018, Hanafy2018, Ghosal2018, Saleena2018, Yan2017, Tong2017, Dedhia2017, Wijayanti2017, Xia2017, Jianqiang2016, Prusa2015, Fersini2015, Abdelwahab2015, Kanakaraj2015, Hagen2015, Cai2015, Mansour2015, WangMingqiang2014, Tsakalidis2014, DaSilva2014, Hassan2013, Goncalves2013} used ensemble learning methods in their work, where they combined the output of several base machine learning and/or deep learning methods. In particular, \cite{Goncalves2013} compared eight popular lexicon and machine learning-based sentiment analysis algorithms, and then developed an ensemble that combines them, which in turn provided the best coverage results and competitive agreement. Moreover, \cite{Ghosal2018} proposes an MLP-based ensemble network that combines LSTM, CNN and feature-based MLP models, with each model incorporating character, word and lexicon level information, to predict the degree of intensity for sentiment and emotion.
Lastly, as presented in Table \ref{table:studies-hybrid-ml}, the RF ensemble learning method was used in the 21 studies \cite{DaSilva2014, Porshnev2014b, Samoylov2014, Yuan2014, Buddhitha2015, Kanakaraj2015, Jianqiang2015b, Bouchlaghem2016, Deshwal2016, Jianqiang2016, Yan2016, Tong2017, Jianqiang2017, Bouazizi2017, Elouardighi2017, Bouazizi2018, LiYujiao2018, Saleena2018, Villegas2018, Yan2018, ZhangYazhou2018}.

\subsubsection{Other}
\label{sssec_other_approaches}
%manual or N/A

In total, 23 studies did not adopt any of the previous approaches (discussed in Sections \ref{sssec_lexicon_approaches}-\ref{sssec_hybrid_approaches}). This is mainly due to three reasons: no information provided by the authors (13 studies), use of an automated approach (4 studies), or use of a manual approach (6 studies) \cite{Sandoval2018, Fang2017, Song2017, Zafar2016, Furini2016, Cvijikj2011} to perform a form of SOM. Regarding the former, the majority of them \cite{Ayoub2018, Tiwari2017, Ouyang2017, Anggoro2016, Williamson2016, Agrawal2014, Pupi2014, Das2014} were not specifically focused on SOM (this was secondary), in contrast to the others \cite{Vivanco2017, Gonzalez2017, Chen2016, Barapatre2016, Mejova2012}. As for the automated approaches \cite{Sharma2018, PaiRajesh2018, Ali2018, Teixeira2017}, some of them used cloud services, such as Microsoft Azure Text Analytics\footnote{https://azure.microsoft.com/en-us/services/cognitive-services/text-analytics/} or out-of-the-box functionality provided by existing tools/software libraries, such as the TextBlob\footnote{https://textblob.readthedocs.io/en/dev/} Python library.

\subsection{Social Datasets}
\label{ssec_social_datasets}

Numerous datasets were used across the 465 studies evaluated for this systematic review. These consisted of SOM datasets released online for public use --which have been widely used across the studies-- and newly collected datasets, some of which were made available for public use or else for private use within the respective studies. In terms of data collection, the majority of them used the respective platform's API, such as the Twitter Search API\footnote{https://developer.twitter.com/en/docs/tweets/search/overview}, either directly or through a third-party library, e.g., Twitter4J\footnote{http://twitter4j.org/en/ - a Java library for the Twitter API}. Due to the large number of datasets, only the ones mostly used shall be discussed within this section. In addition, only social datasets are mentioned irrespective of whether other non-social datasets (e.g., news, movies, etc.,) were used, given that the main focus of this review is on social data.

The first sub-section (Section \ref{sssec_social_datasets_overview}) presents an overview of the top social datasets used, whereas the second sub-section (Section \ref{sssec_social_datasets_comparative}) presents a comparative analysis of the studies that produced the best performance for each respective social dataset.

\subsubsection{Overview}
\label{sssec_social_datasets_overview}

The following are the top ten social datasets used across all studies:

\begin{enumerate}
	\item \textbf{Stanford Twitter Sentiment (STS)} \cite{Go2009} used in 61 studies: 1,600,000 training tweets collected via the Twitter API, that is made up of 800,000 tweets containing positive emoticons and 800,000 tweets containing negative emoticons. These are based on various topics, such as Nike, Google, China, Obama, Kindle, San Francisco, North Korea and Iran.  
	\item \textbf{Sanders\footnote{original dataset points to \url{http://www.sananalytics.com/lab/twitter-sentiment/} which is not online anymore}} - used in 32 studies: 5513 hand-classified tweets about four topics: Apple, Google, Microsoft, Twitter. These tweets are labelled as follows: 570 positive, 654 negative, 2503 neutral, and 1786 irrelevant. 
	\item \textbf{SemEval 2013 - Task 2\footnote{https://www.cs.york.ac.uk/semeval-2013/task2/}} \cite{Nakov2013} - used in 28 studies: Training, development and test sets for Twitter and SMS messages were annotated with positive, negative, and objective/neutral labels via the Amazon Mechanical Turk crowdsourcing platform. This was done for 2 subtasks focusing on an expression-level and message-level. 	
	\item \textbf{SemEval 2014 - Task 9\footnote{http://alt.qcri.org/semeval2014/task9/}} \cite{Rosenthal2014} - used in 18 studies: Continuation of SemEval 2013 - Task 2, where three new test sets from regular and sarcastic tweets, and LiveJournal sentences were introduced.
	\item \textbf{STS Gold (STS-Gold)} \cite{Saif2013} - used in 17 studies: A subset of STS, which was annotated manually at a tweet and entity-level. The tweet labels were either positive, negative, neutral, mixed, or other.  
	\item \textbf{Health care reform (HCR)} \cite{Speriosu2011} - used in 17 studies: Dataset contains tweets about the 2010 health care reform in the USA. A subset of these are annotated for polarity with the following labels: positive, negative, neutral, irrelevant. The polarity targets, such as health care reform, conservatives, democrats, liberals, republicans, Obama, Stupak and Tea Party, were also annotated. All were distributed into training, development and test sets.
	\item \textbf{Obama-McCain Debate (OMD)} \cite{Shamma2009} - used in 17 studies: 3,238 tweets about the first presidential debate held in the USA for the 2008 campaign. The sentiment labels of the tweets are acquired by \cite{Diakopoulos2010} using Amazon Mechanical Turk, and are rated as either positive, negative, mixed, or other. 
	\item \textbf{SemEval 2015 - Task 10\footnote{http://alt.qcri.org/semeval2015/task10/}} \cite{Rosenthal2015} - used in 15 studies: This continues on datasets number 3 and 4, with three new subtasks. The first two target sentiment about a particular topic in one tweet or collection of tweets, whereas the third targets the degree of prior polarity of a phrase. 	
	\item \textbf{SentiStrength Twitter (SS-Twitter)} \cite{Thelwall2012} - used in 12 studies: 6 human-coded databases from BBC, Digg, MySpace, Runners World, Twitter and YouTube annotated for sentiment polarity strength i.e., negative between -1 (not negative) and -5 (extremely negative), and positive between 1 (not positive) and 5 (extremely positive). 
	\item \textbf{SemEval 2016 - Task 4\footnote{http://alt.qcri.org/semeval2016/task4/}} \cite{Nakov2016} - used in 9 studies: This is a re-run of dataset 7, with three new subtasks. The first one replaces the standard two-point scale (positive/negative) or three-point scale (positive/negative/neutral) with a five-point scale (very positive/positive/OK/ negative/very negative). The other two subtasks replaced tweet classification with quantification (i.e., estimating the distribution of the classes in a set of unlabelled items) according to a two-point and five-point scale, respectively. 
	\item \textbf{NLPCC 2012\footnote{http://tcci.ccf.org.cn/conference/2012/}} - used in 6 studies: Chinese microblog sentiment dataset (sentence level) from Tencent Weibo provided by the First Conference on Natural Language Processing and Chinese Computing (NLP\&CC 2012) It consists of a training set of microblogs about two topics, and a test set about 20 topics, where the subjectivity (subjective/objective) and the polarity (positive/negative/neutral) was assigned for each.
	\item \textbf{NLPCC 2013\footnote{http://tcci.ccf.org.cn/conference/2013/}} - used in 6 studies: Dataset from Sina Weibo used for the Chinese Microblog Sentiment Analysis Evaluation (CMSAE) task in the second conference on NLP\&CC 2013. The Chinese microblogs were classified into 7 emotion types: anger, disgust, fear, happiness, like, sadness, surprise. Test set contains 10,000 microblogs, where each text is labelled with a primary emotion type ans a secondary one (if possible).	
	\item \textbf{Sentiment Evaluation (SE-Twitter)} \cite{Narr2012} - used in 5 studies: Human annotated multilingual dataset of 12,597 tweets from 4 languages, namely English, German, French, Portuguese. Polarity annotations with labels: positive, negative, neutral, and irrelevant, were conducted manually using Amazon Mechanical Turk.  
	\item \textbf{SemEval 2017 - Task 4} \cite{Rosenthal2017} - used in 5 studies: This dataset continues with a re-run of dataset 10, where two new changes were introduced; inclusion of the Arabic language for all subtasks and provision of profile information of the Twitter users that posted the target tweets.
\end{enumerate}	

All the datasets above are textual, with the majority of them composed of social data from Twitter. From the datasets above, in terms of language, only the SE-Twitter (number 13) social dataset can be considered as multilingual, with the rest targeting English (majority) or Chinese microblogs, whereas SemEval 2017 - Task 4 (number 14) introduced a new language in Arabic. An additional dataset is the one produced by Mozeti{\v c} et al., which contains 15 Twitter sentiment corpora for 15 European languages \cite{Mozetic2016}. Some studies such as \cite{Munezero2015} used one of the English-based datasets above (STS-Gold) for multiple languages, given that they adopted a lexicon-based approach. Moreover, these datasets had different usage within the respective studies, with the most common being used as a training/test set, the final evaluation of the proposed solution/lexicon, or for comparison purposes. Evaluation challenges like SemEval are important to generate social datasets such as the above and \cite{Cortis2017}, since these can be used by the Opinion Mining community for further research and development.

\subsubsection{Comparative Analysis}
\label{sssec_social_datasets_comparative}

A comparative analysis of all the studies that used the social datasets presented in the previous sub-section was carried out. The Precision, Recall, F-measure (F1-score), and Accuracy metrics were selected to evaluate the said studies (when available) and identify the best performance for each respective social dataset. It is important to note that for certain datasets, this could not be done, since the experiments conducted were not consistent across all the studies. The top three studies (where possible) obtaining the best results for each of the four evaluation metrics are presented in the tables below. 

Tables \ref{table:dataset-STS} and \ref{table:dataset-Sanders} provide the best results for the STS and Sanders datasets.
\begin{table} [h!]
    \small
	\begin{center}
    	\begin{tabular}{| l | l | l | l | l |}
    	\hline
     	\textbf{Ranking} & \textbf{Precision} & \textbf{Recall} & \textbf{F-measure} & \textbf{Accuracy}  \\ \hline
		1 & 87.60\% \cite{Jianqiang2018}  &	91.76\% \cite{Siddiqua2016}  & 87.50\% \cite{Jianqiang2018} & 89.61\% \cite{Shyamasundar2016}  	  	\\ \hline    
		2 & 85.00\% \cite{Ismail2018} &	87.60\% \cite{Bravo2013} & 86.08\% \cite{Siddiqua2016} & 88.80\% \cite{Arslan2018}  	  		\\ \hline        
     	3 & 84.56\% \cite{Zainuddin2016a} &	87.45\% \cite{Jianqiang2018} & 83.90\% \cite{Saif2012} & 88.30\% \cite{Lek2013}		     	\\ \hline              
    	\end{tabular}
	\end{center}
	\caption{Studies obtaining the best performance for the STS (1) social dataset}
	\label{table:dataset-STS}
\end{table}

\begin{table} [h!]
    \small
	\begin{center}
    	\begin{tabular}{| l | l | l | l | l |}
    	\hline
     	\textbf{Ranking} & \textbf{Precision} & \textbf{Recall} & \textbf{F-measure} & \textbf{Accuracy}  \\ \hline
		1 & 97.72\% \cite{Ameur2018} & 97.41\% \cite{Ameur2018}	& 98.20\% \cite{Bravo2014} & 98.10\% \cite{Bravo2014} \\ \hline    
		2 &	79.00\% \cite{Bravo2013} & 89.10\% \cite{Bravo2013} & 97.57\% \cite{Ameur2018} & 88.93\% \cite{Korenek2014} \\ \hline        
     	3 & 77.60\% \cite{Deshwal2016} & 78.70\% \cite{Deshwal2016}	& 84.85\% \cite{DaSilva2014} & 88.30\% \cite{Celiktug2018} \\ \hline                            
		\end{tabular}
	\end{center}
	\caption{Studies obtaining the best performance for the Sanders (2) social dataset}
	\label{table:dataset-Sanders}
\end{table}

\newpage 
Tables \ref{table:dataset-SemEval2013T2} and \ref{table:dataset-SemEval2014T9} provide the best results for the SemEval 2013 - Task 2 and SemEval 2014 - Task 9 datasets, specifically for sub-task B, which focused on message polarity classification. Moreover, the results obtained by the participants of this shared task should be reviewed for a more representative comparative evaluation. 

\begin{table} [h!]
    \small
	\begin{center}
    	\begin{tabular}{| l | l | l | l | l |}
    	\hline
     	\textbf{Ranking} & \textbf{Precision} & \textbf{Recall} & \textbf{F-measure} & \textbf{Accuracy}  \\ \hline
		1 & 80.69\% \cite{Chikersal2015} & 83.68\% \cite{Chikersal2015} & 93.70\% \cite{Bravo2014} & 93.70\% \cite{Bravo2014} \\ \hline    
		2 &	NA & NA	& 81.90\% \cite{Chikersal2015} & 91.16\% \cite{Lima2015} \\ \hline        
     	3 & NA & NA	& 80.30\% \cite{Xia2017} & 89.00\% \cite{Yan2018} \\ \hline                            
		\end{tabular}
	\end{center}
	\caption{Studies obtaining the best performance for the SemEval 2013 - Task 2 (3) social dataset}
	\label{table:dataset-SemEval2013T2}
\end{table}
   
\begin{table} [h!]
    \small
	\begin{center}
    	\begin{tabular}{| l | l | l | l | l |}
    	\hline
     	\textbf{Ranking} & \textbf{Precision} & \textbf{Recall} & \textbf{F-measure} & \textbf{Accuracy}  \\ \hline
		1 & 83.56\% \cite{Jianqiang2018} & 81.48\% \cite{Jianqiang2018} & 82.36\% \cite{Xia2017} & 85.82\% \cite{Jianqiang2018} \\ \hline    
		2 & 80.47\% \cite{Jianqiang2016} & 80.98\% \cite{Chikersal2015} & 81.99\% \cite{Jianqiang2018} & 83.82\% \cite{Jianqiang2016} \\ \hline        
     	3 & 78.93\% \cite{Chikersal2015} & 76.89\% \cite{Jianqiang2016}	& 79.81\% \cite{Chikersal2015} & 83.06\% \cite{Jianqiang2015a} \\ \hline                            
		\end{tabular}
	\end{center}
	\caption{Studies obtaining the best performance for the SemEval 2014 - Task 9 (4) social dataset}
	\label{table:dataset-SemEval2014T9}
\end{table}

Tables \ref{table:dataset-STSGold}, \ref{table:dataset-HCR} and \ref{table:dataset-OMD} provide the best results for the STS-Gold, HCR and OMD datasets.

\begin{table} [h!]
    \small
	\begin{center}
    	\begin{tabular}{| l | l | l | l | l |}
    	\hline
     	\textbf{Ranking} & \textbf{Precision} & \textbf{Recall} & \textbf{F-measure} & \textbf{Accuracy}  \\ \hline
		1 & 82.75\% \cite{Jianqiang2018} & 82.61\% \cite{Jianqiang2018}	& 83.10\% \cite{Saif2014b} & 92.67\% \cite{Krouska2016} \\ \hline    
		2 &	82.20\%	\cite{Ismail2018} & 82.30\% \cite{Ismail2018} & 82.65\% \cite{Jianqiang2018} & 89.02\% \cite{Troussas2016} \\ \hline        
        3 & 79.26\% \cite{Saif2014a} & 80.04\% \cite{Saif2014a}	& 79.62\% \cite{Saif2014a} & 86.37\% \cite{Yan2016} \\ \hline                            
		\end{tabular}
	\end{center}
	\caption{Studies obtaining the best performance for the STS-Gold (5) social dataset}
	\label{table:dataset-STSGold}
\end{table}

%Table \ref{table:dataset-HCR} provides the best results for the HCR dataset.

\begin{table} [h!]
    \small
	\begin{center}
    	\begin{tabular}{| l | l | l | l | l |}
    	\hline
     	\textbf{Ranking} & \textbf{Precision} & \textbf{Recall} & \textbf{F-measure} & \textbf{Accuracy}  \\ \hline
		1 & 71.30\% \cite{Mishra2018} & 67.40\% \cite{Saif2012}	& 70.28\% \cite{Saleena2018} & 91.94\% \cite{Krouska2016} \\ \hline    
		2 & 69.15\% \cite{Saif2012}	& 59.47\% \cite{Saif2014a} & 69.10\% \cite{Saif2014b} & 85.10\% \cite{Troussas2016} \\ \hline        
     	3 & 59.76\% \cite{Saif2014a} & 58.30\% \cite{Mishra2018} & 68.00\% \cite{DaSilva2014} & 84.50\% \cite{Yan2018} \\ \hline                            
		\end{tabular}
	\end{center}
	\caption{Studies obtaining the best performance for the HCR (6) social dataset}
	\label{table:dataset-HCR}
\end{table}

\begin{table} [h!]
    \small
	\begin{center}
    	\begin{tabular}{| l | l | l | l | l |}
    	\hline
     	\textbf{Ranking} & \textbf{Precision} & \textbf{Recall} & \textbf{F-measure} & \textbf{Accuracy}  \\ \hline
		1 & 81.36\% \cite{ZhangYazhou2018} & 79.00\% \cite{Saif2012} & 81.34\% \cite{Saif2014b}	& 92.59\% \cite{Krouska2016} \\ \hline    
		2 &	78.95\% \cite{Saif2012}	& 65.76\% \cite{Saif2014a} & 78.20\% \cite{Saif2012} & 87.74\% \cite{Troussas2016} \\ \hline        
     	3 & 66.51\% \cite{Saif2014a} & 61.60\% \cite{Mishra2018} & 74.65\% \cite{DaSilva2014} & 82.90\% \cite{Saif2014b} \\ \hline                            
		\end{tabular}
	\end{center}
	\caption{Studies obtaining the best performance for the OMD (7) social dataset}
	\label{table:dataset-OMD}
\end{table}

\newpage 
Table \ref{table:dataset-SemEval2015T10} provides the best results for the SemEval 2015 - Task 10 dataset, specifically for sub-task B, which focused on message polarity classification. Moreover, the results obtained by the participants of this shared task should be reviewed for a more representative comparative evaluation. 
\begin{table} [h!]
    \small
	\begin{center}
    	\begin{tabular}{| l | l | l | l | l |}
    	\hline
     	\textbf{Ranking} & \textbf{Precision} & \textbf{Recall} & \textbf{F-measure} & \textbf{Accuracy}  \\ \hline
		1 & NA & NA	& 76.02\% \cite{Xia2017} & 68.77\% \cite{Stojanovski2015} \\ \hline    
		2 &	NA & NA	& 67.39\% \cite{Sygkounas2016} & 68.00\% \cite{Li2017} \\ \hline        
     	3 & NA & NA	& 64.88\% \cite{Stojanovski2018} & 51.95\% \cite{Stojanovski2018} \\ \hline                            
		\end{tabular}
	\end{center}
	\caption{Studies obtaining the best performance for the SemEval 2015 - Task 10 (8) social dataset}
	\label{table:dataset-SemEval2015T10}
\end{table}

Table \ref{table:dataset-SSTwitter} provides the best results for the SS-Twitter dataset.        

\begin{table} [h!]
    \small
	\begin{center}
    	\begin{tabular}{| l | l | l | l | l |}
    	\hline
     	\textbf{Ranking} & \textbf{Precision} & \textbf{Recall} & \textbf{F-measure} & \textbf{Accuracy}  \\ \hline
		1 & 80.61\% \cite{Jianqiang2018} & 80.86\% \cite{Jianqiang2018}	& 80.72\% \cite{Jianqiang2018} & 89.10\% \cite{Yan2018} \\ \hline    
		2 &	67.77\% \cite{ZhangYazhou2018} & 54.77\% \cite{ZhangYazhou2018}	& 72.27\% \cite{Saif2014b} & 84.59\% \cite{Lima2015} \\ \hline        
     	3 & NA & NA	& 59.27\% \cite{ZhangYazhou2018} & 81.56\% \cite{Su2017} \\ \hline                            
		\end{tabular}
	\end{center}
	\caption{Studies obtaining the best performance for the SS-Twitter (9) social dataset}
	\label{table:dataset-SSTwitter}
\end{table}    

Table \ref{table:dataset-SemEval2016T4} provides the best results for the SemEval 2016 - Task 4 dataset, specifically for sub-task A, which focused on message polarity classification. Moreover, the results obtained by the participants of this shared task should be reviewed for a more representative comparative evaluation. 
     	                           	
\begin{table} [h!]
    \small
	\begin{center}
    	\begin{tabular}{| l | l | l | l | l |}
    	\hline
     	\textbf{Ranking} & \textbf{Precision} & \textbf{Recall} & \textbf{F-measure} & \textbf{Accuracy}  \\ \hline
		1 & 64.10\% \cite{Mishra2018} & 60.50\% \cite{Mishra2018} & 77.25\% \cite{Xia2017} & 65.60\% \cite{Mishra2018} \\ \hline    
		2 & NA & NA	& 61.40\% \cite{Mishra2018} & NA \\ \hline        
     	3 & NA & NA	& 57.10\% \cite{Villegas2018} & NA \\ \hline                            
		\end{tabular}
	\end{center}
	\caption{Studies obtaining the best performance for the SemEval 2016 - Task 4 (10) social dataset}
	\label{table:dataset-SemEval2016T4}
\end{table}     

Tables \ref{table:dataset-NLPCC2012T1} and \ref{table:dataset-NLPCC2012T2} provide the best results for the NLPCC 2012 dataset. Results quoted below are for task 1 which focused on subjectivity classification (see Table \ref{table:dataset-NLPCC2012T1}) and task 2 which focused on sentiment polarity classification (see Table \ref{table:dataset-NLPCC2012T2}). Moreover, the results obtained by the participants of this shared task should be reviewed for a more representative comparative evaluation.   	

\begin{table} [h!]
    \small
	\begin{center}
    	\begin{tabular}{| l | l | l | l | l |}
    	\hline
     	\textbf{Ranking} & \textbf{Precision} & \textbf{Recall} & \textbf{F-measure} & \textbf{Accuracy}  \\ \hline
		1 & 72.20\% \cite{Cui2013} & 96.70\% \cite{Feng2015} & 78.80\% \cite{Feng2015} &  NA \\ \hline    
		2 &	69.15\% \cite{Hao2017} & 96.00\% \cite{Shi2013}  & 77.00\% \cite{Shi2013}  &  NA \\ \hline        
     	3 & 66.90\% \cite{Feng2015}& 73.80\% \cite{Cui2013}  & 72.10\% \cite{Cui2013}  &  NA \\ \hline                            
		\end{tabular}
	\end{center}
	\caption{Studies obtaining the best performance for the NLPCC 2012 - Task 1 (11) social dataset}
	\label{table:dataset-NLPCC2012T1}
\end{table}    

\begin{table} [h!]
    \small
	\begin{center}
    	\begin{tabular}{| l | l | l | l | l |}
    	\hline
     	\textbf{Ranking} & \textbf{Precision} & \textbf{Recall} & \textbf{F-measure} & \textbf{Accuracy}  \\ \hline
		1 & 80.90\% \cite{Shi2013} & 77.80\% \cite{Shi2013}  & 79.30\% \cite{Shi2013} &  NA \\ \hline    
		2 &	78.60\% \cite{Cui2013} & 74.60\% \cite{Feng2015} & 69.14\% \cite{Hao2017} &  NA \\ \hline        
     	3 & 70.83\% \cite{Hao2017} & 67.52\% \cite{Hao2017}  & 67.10\% \cite{Cui2013} &  NA \\ \hline                            
		\end{tabular}
	\end{center}
	\caption{Studies obtaining the best performance for the NLPCC 2012 - Task 2 (11) social dataset}
	\label{table:dataset-NLPCC2012T2}
\end{table}     

\newpage 
Tables \ref{table:dataset-NLPCC2013} and \ref{table:dataset-SETwitter} provide the best results for the NLPCC 2013 and SE-Twitter datasets.                          
     
\begin{table} [h!]
    \small
	\begin{center}
    	\begin{tabular}{| l | l | l | l | l |}
    	\hline
     	\textbf{Ranking} & \textbf{Precision} & \textbf{Recall} & \textbf{F-measure} & \textbf{Accuracy}  \\ \hline
		1 & NA & NA	& 83.02\% \cite{Xia2017} & 78.80\% \cite{Jiang2015} \\ \hline    
		2 &	NA & NA	& NA 					 & 63.90\% \cite{Jiang2013} \\ \hline        
     	3 & NA & NA	& NA 					 & NA \\ \hline                            
		\end{tabular}
	\end{center}
	\caption{Studies obtaining the best performance for the NLPCC 2013 (12) social dataset}
	\label{table:dataset-NLPCC2013}
\end{table}   

%Table \ref{table:dataset-SETwitter} provides the best results for the SE-Twitter dataset.                          

\begin{table} [h!]
    \small
	\begin{center}
    	\begin{tabular}{| l | l | l | l | l |}
    	\hline
     	\textbf{Ranking} & \textbf{Precision} & \textbf{Recall} & \textbf{F-measure} & \textbf{Accuracy}  \\ \hline
		1 & 88.00\% \cite{Jianqiang2018} & 87.32\% \cite{Jianqiang2018}	& 87.66\% \cite{Jianqiang2018} &   87.39\% \cite{Jianqiang2018} \\ \hline    
		2 &	86.16\% \cite{Jianqiang2016} & 86.15\% \cite{Jianqiang2016}	& 86.08\% \cite{Jianqiang2016} &  86.72\% \cite{Jianqiang2016} \\ \hline        
     	3 & NA						     & NA					 		& NA						   &  85.87\% \cite{Jianqiang2015a} \\ \hline                            
		\end{tabular}
	\end{center}
	\caption{Studies obtaining the best performance for the SE-Twitter (13) social dataset}
	\label{table:dataset-SETwitter}
\end{table}      

\newpage 
Table \ref{table:dataset-SemEval2017T4} provides the best results for the SemEval 2017 - Task 4 dataset, specifically for sub-task A, which focused on message polarity classification. Moreover, the results obtained by the participants of this shared task should be reviewed for a more representative comparative evaluation.     
     	     
\begin{table} [h!]
    \small
	\begin{center}
    	\begin{tabular}{| l | l | l | l | l |}
    	\hline
     	\textbf{Ranking} & \textbf{Precision} & \textbf{Recall} & \textbf{F-measure} & \textbf{Accuracy}  \\ \hline
		1 & NA & NA	& 62.30\% \cite{Villegas2018} & 67.30\% \cite{Symeonidis2018} \\ \hline    
		2 &	NA & NA	& NA 						  & 60.70\% \cite{Effrosynidis2017} \\ \hline        
     	3 & NA & NA	& NA 						  & NA \\ \hline                            
		\end{tabular}
	\end{center}
	\caption{Studies obtaining the best performance for the SemEval 2017 - Task 4 (14) social dataset}
	\label{table:dataset-SemEval2017T4}
\end{table}

The following are some comments regarding the social dataset results quoted in the tables above:
\begin{itemize}
	\item In cases where several techniques and/or methods were applied, the highest result obtained in the study for each of the four evaluation metrics, was recorded, even if the technique did not produce the best result for all metrics.	
	\item The average Precision, Recall, and F-measure results are quoted (if provided by authors), i.e., average score of the results for each classified level (e.g., the average score of the results obtained for each sentiment polarity classification level - positive, negative and, neutral).
	\item Results for social datasets that were released as a shared evaluation task, such as SemEval, were either only provided in the metrics used by the task organisers or other metrics were chosen by the authors, therefore not quoted.	
	\item Certain studies evaluated their techniques based on a subset of the actual dataset. Results quoted are the ones where the entire dataset was used (according to the authors and/our our understanding).	
	\item Quoted results are for classification tasks and not aspect-based SOM, which can vary depending on the focus of the study.	
	\item Results presented in a graph visualisation were not considered due to the exact values not being clear.		
\end{itemize}

\subsection{Language}
\label{ssec_language}

Multilingual/bilingual SOM is very challenging, since it deals with multi-cultural social data. For example, analysing Chinese and English online posts can bring a mixed sentiment on such posts. Therefore, it is hard for researchers to make a fair judgement in cases where online posts' results from different languages contradict each other \cite{YanG2014}. 

The majority of the studies (354 out of 465) considered for this review analysis support one language in their SOM solutions. A total of 80 studies did not specify whether their proposed solution is language-agnostic or otherwise, or else their modality was not textual-based. Lastly, only 31 studies cater for more than one language, with 18 being bilingual, 1 being trilingual and 12 proposed solutions claiming to be multilingual. Regarding the latter, the majority were tested on a few languages at most, with \cite{Castellucci2015a, Castellucci2015b} on English and Italian, \cite{Montejo2014} on English and Spanish, \cite{Erdmann2014} on English and Japanese, \cite{Radhika2017} on English and Malayalam\footnote{Malayalam is a Dravidian language spoken in the Indian state of Kerala and the union territories of Lakshadweep and Puducherry by the Malayali people}, \cite{Baccouche2018} on English, French and Arabic, \cite{Munezero2015} on keyword sets for different languages (e.g., Spanish, French), \cite{Wehrmann2017} on English, Spanish, Portuguese and German, \cite{Cui2011} on Basic Latin (English) and Extended Latin (Portuguese, Spanish, German), \cite{Teixeira2017} on  Spanish, Italian, Portuguese, French, English, and Arabic, \cite{Zhang2017} on 8 languages, namely English, German, Portuguese, Spanish, Polish, Slovak, Slovenian, Swedish, and \cite{GaoBo2016} on 11 languages, namely English, Dutch, French, German, Italian, Polish, Portuguese, Russian, Spanish, Swedish and Turkish. 

The list below specifies the languages supported by the 19 bilingual and trilingual studies:
\begin{multicols}{2}
	\begin{itemize}
		\item English and Italian \cite{Severyn2016, Davanzo2015, Pupi2014};
		\item English and German \cite{Abdelrazeq2016, Tumasjan2010};
		\item English and Spanish \cite{Giachanou2017, Cotfas2015, Delcea2014};
		\item English and Brazilian Portuguese \cite{Guerra2014};
		\item English and Chinese \cite{Xia2017, YanG2014};
		\item English and Dutch \cite{Flaes2016};
		\item English and Greek \cite{Politopoulou2013};
		\item English and Hindi \cite{Anjaria2014};
		\item English and Japanese \cite{Ragavi2014};
		\item English and Roman-Urdu \cite{Javed2014};
		\item English and Swedish \cite{LiYujiao2018};
		\item English and Korean \cite{Ramadhani2017};
		\item English, German and Spanish \cite{Boididou2018}.
	\end{itemize}
\end{multicols}

Some studies above \cite{Davanzo2015, Anjaria2014, Tumasjan2010} translated their input data into an intermediate language, mostly English, to perform SOM. 

%\vspace*{-7mm}
Moreover, Table \ref{table:studies-language} provides a list of the non-English languages identified from the 354 studies that support one language. 
Authors in \cite{Chou2017} claim that their method can be easily applied to any ConceptNet\footnote{http://conceptnet.io/ -- an open, multilingual knowledge graph} supported language, with \cite{WangYang2016} similarly claiming that their method is language independent, whereas the solution by \cite{Wang2015} is multilingual given that emoticons are used in the majority of languages.

\begin{table} [h!]
    \small
	\begin{center}
    	\begin{tabular}{ | p{3.2cm} | p{0.8cm} | p{6.5cm} |}
    	\hline
     	           \textbf{Language}           & \textbf{Total} & \textbf{Studies} \\ \hline
     	            
     	           Chinese            & 53    & \cite{Cao2018, LiD2018, LiuQuanchao2018, SunXiaoZhang2018, WangHongwei2018, Wan2018, Chou2017, Hao2017, Ouyang2017, Shi2017, Sun2017, Zhang2017, ZhangL2013, GaoFei2016, Liu2016, Zhao2016, WangYang2016, Wu2016c, LiW2016, YangY2015, ChenF2015, WangZhitao2014, Sui2012, Yanmei2015, Liu2015, Zhang2015, WangMin2014, Tian2015, Feng2015, Song2015, Jiang2015, Kuo2016, WangYaqi2016, Wang2015, Du2014, Gao2015, ChenP2015, WangMingqiang2014, SuZ2014, Ou2014, YanB2014, Pei2014, Sun2014, WangDongfang2014, Xiong2013, Zhu2013, Jiang2013, ZhangH2013, Tang2013, Cui2013, Shi2013, Zhang2012, Li2014}  \\ \hline
     	           Spanish            & 11     &  \cite{Calvo2018, Hubert2018, Ochoa2018, Sanchez2018, Gonzalez2017, Tellez2017, Gambino2016, Tapia2014, Molina2014, Pla2014, Ortigosa2014} \\ \hline
     	           Indonesian         & 8     & \cite{Fitri2018, Permatasari2018, Setiawan2018, Napitu2017, Nugroho2017, Rinaldi2017, Wijayanti2017, Ramadhani2016} \\ \hline
                   Italian            & 5     & \cite{Buscaldi2015, Furini2016, Santarcangelo2015, Bosco2013, Hernandez2014} \\ \hline
                   Arabic 			  & 5     & \cite{Alzahrani2018, Elouardighi2017, Bouchlaghem2016, Khalil2015, Abdul2014}   \\ \hline
                   Portuguese 		  & 3     & \cite{Kokkinogenis2015, Seron2015, Souza2012} \\ \hline
     	           Brazilian Portuguese & 3     & \cite{DosSantos2018, Souza2016, Costa2014} \\ \hline
     	           Japanese           & 3     & \cite{Tago2018, Ishikawa2017, Maeda2012}        \\ \hline
     	           Korean             & 2     & \cite{Cho2014, Park2011} \\ \hline
     	           French 			  & 2     & \cite{Ameur2018, Lai2015} \\ \hline
     	           French - Bambara	  & 1 	  & \cite{Konate2018} \\ \hline
     	           Bulgarian          & 1     & \cite{Smailovic2015} \\ \hline       	
     	           German             & 1     & \cite{Rill2014} \\ \hline
     	           Roman Urdu         & 1     & \cite{Zafar2016} \\ \hline     	              
     	           Russian			  & 1     & \cite{Averchenkov2015} \\ \hline
     	           Swiss German       & 1     & \cite{Cvijikj2011} \\ \hline
     	           Thai				  & 1     & \cite{Wunnasri2013} \\ \hline         
     	           Persian			  & 1	  & \cite{Salari2018} \\ \hline 
     	           Bengala			  & 1	  & \cite{Sabuj2017} \\ \hline
     	           Vietnamese 		  & 1 	  & \cite{Vo2017} \\ \hline         
    	\end{tabular}
	\end{center}
	\caption{Non-English languages supported by studies in this review analysis}
	\label{table:studies-language}
\end{table}

\subsection{Modality}
\label{ssec_modality}

The majority of the studies in this systematic review and in the state-of-the-art focus on SOM on the textual modality, with only 15 out of 465 studies applying their work on more than one modality. However, other modalities, such as visual (image, video), and audio information is often ignored, even though it contributes greatly towards expressing user emotions \cite{ChenF2015}. Moreover, when two or more modalities are considered together for any form of social opinion, such as emotion recognition, they are often complementary, thus increase the system's performance \cite{Caschera2016}. Table \ref{table:studies-multimodal} lists the multimodal studies within the review analysis, with the ones catering for two modalities --text and image-- being the most popular. 

%\vspace*{-4mm}

\begin{table} [h!]
	\small
	\begin{center}
    	%\begin{tabular}{ | c | c | c | c | c |}
    	\begin{tabular}{ | p{1cm} | p{1cm} | p{1cm} | p{1cm} | p{6.25cm} |}
    	\hline
     	             \textbf{Text} & \textbf{Image}  & \textbf{Video}  & \textbf{Audio}  & \textbf{Studies} \\ \hline
     	           \cmark & \cmark &        &        & \cite{Ortis2018, Rai2018, Saini2018, Chen2017, ChenF2015, Baecchi2016, Liu2015, Zhang2015, WangMin2014, Flaes2016, Cai2015, Yuan2015} \\ \hline
     	                  &        & \cmark & \cmark & \cite{Song2017} \\ \hline
     	           \cmark &        & \cmark & \cmark & \cite{Caschera2016} \\ \hline                                            	               \cmark & \cmark & \cmark & \cmark & \cite{Poria2016} \\ \hline   	       
    	\end{tabular}
	\end{center}
	\caption{Studies adopting a multimodal approach}
	\label{table:studies-multimodal}
\end{table}

%\vspace*{-6mm}

\subsubsection{Datasets}

Current available datasets and resources for SOM are restricted to the textual modality only. The following are the non-textual social datasets (not listed in Section \ref{ssec_social_datasets}) used across the mentioned studies:
\begin{itemize}
	\item \textbf{YouTube Dataset} \cite{Morency2011} used in \cite{Poria2016}: 47 videos targeting various topics, such as politics, electronics and product reviews.
	\item \textbf{SentiBank Twitter Dataset\footnote{http://www.ee.columbia.edu/ln/dvmm/vso/download/sentibank.html}} \cite{Borth2013} used in \cite{Baecchi2016, Cai2015}: Image dataset from Twitter annotated for polarity using Amazon Mechanical Turk.
Tweets with images related to 21 hashtags (topics) resulted in 470 being positive and 133 being negative. 
	\item \textbf{SentiBank Flickr Dataset} \cite{Borth2013} used in \cite{Cai2015}: 500,000 image posts from Flickr labeled by 1,553 adjective noun pairs based on Plutchik's Wheel of Emotions (psychological theory) \cite{Plutchik1980}.
	\item \textbf{You Image Dataset} \cite{You2015} used in \cite{Cai2015}:
Image dataset from Twitter consisting of 769 positive and 500 negative tweets with images, annotated using Amazon Mechanical Turk. 
	\item \textbf{Katsurai and Sotoh Image Dataset\footnote{http://mm.doshisha.ac.jp/senti/CrossSentiment.html}} \cite{Katsurai2016} used in \cite{Ortis2018}: Dataset of images from Flickr (90,139) and Instagram (65,439) with their sentiment labels.	
\end{itemize}

\subsubsection{Observations}

The novel methodology by Poria et al. \cite{Poria2016}, is the only mutlimodal sentiment analysis approach which caters for four different modalities, namely text, vision (image and video) and audio. Sentiments are extracted from social Web videos. In \cite{Caschera2016}, the authors propose a method whereby machine learning techniques need to be trained on different and heterogeneous features when used on different modalities, such as polarity and intensity of lexicons from text, prosodic features from audio, and postures, gestures and expressions from video. The sentiment of video and audio data in \cite{Song2017} was manually coded, which task is labour intensive and time consuming. The addition of images to the microblogs' textual data reinforces and clarifies certain feelings \cite{WangMin2014, Baecchi2016}, thus improving the sentiment classifier with the image features \cite{Liu2015, Zhang2015, WangMin2014, Cai2015}. Similarly, \cite{ChenF2015} also demonstrates superiority with their multimodal hypergraph method when compared to single modality (in this case textual) methods. Moreover, these results are further supported by the method in \cite{Poria2016} --which caters for more than two modalities, in audio, visual and textual-- where it shows that accuracy improves drastically when such modalities are used together.  

Flaes et al., \cite{Flaes2016} apply their multimodal (text, images) method in a real world application area, which research shows that several relationships exist between city liveability indicators collected by the local government and sentiment that is extracted automatically. For example, a negative linear association of detected sentiment from Flickr data is related with people living on welfare checks. Results in \cite{Rai2018} show that there is a high correlation between sentiment extracted from text-based social data and image-based landscape preferences by humans. In addition, results in \cite{Yuan2015} show some correlation between image and textual tweets. However, the authors mention that more features and robust data is required to determine the exact influence of multimedia content in the social domain. The work in \cite{Chen2017} adopts a bimodal approach to solve the problem of cross-domain image sentiment classification by using textual features and visual features from the target domain and measuring the text/image similarity simultaneously.

Therefore, multimodality in the SOM domain is one of numerous research gaps identified in this systematic review. This provides researchers with an opportunity towards further research, development and innovation in this area.

\subsection{Tools and Technologies}
\label{ssec_tools}
In this systematic review, we also analysed the tool and technologies that were used across all studies for various opinion mining operations conducted on social data, such as NLP, machine learning, and big data handling. The subsections below provide respective lists for the ones mostly used across the studies for the various operations required.

\subsubsection{NLP}
\label{sssec_tools_nlp}

The following are the top 5 NLP tools used across all studies for various NLP tasks:
\begin{itemize}
	\item Natural Language Toolkit (NLTK)\footnote{https://www.nltk.org/}: a platform that provides lexical resources, text processing libraries for classification, tokenisation, stemming, tagging, parsing, and semantic reasoning, and wrappers for industrial NLP libraries;
	\item TweetNLP\footnote{http://www.cs.cmu.edu/\~ark/TweetNLP/}: consists of a tokeniser, Part-of-Speech (POS) tagger, hierarchical word clusters, and a dependency parser for tweets, besides annotated corpora and web-based annotation tools;
	\item Stanford NLP\footnote{https://nlp.stanford.edu/software/}: software that provides statistical NLP, deep learning NLP and rule-based NLP tools, such as Stanford CoreNLP, Stanford Parser, Stanford POS Tagger;  
	\item NLPIR-ICTCLAS\footnote{http://ictclas.nlpir.org/}: a Chinese word segmentation system that includes keyword extraction, POS tagging, NER, and microblog analysis, amongst other features; 
	\item word2vec\footnote{https://code.google.com/archive/p/word2vec/}: an efficient implementation of the continuous bag-of-words and skip-gram architectures for computing vector representations of words.	
\end{itemize}

\subsubsection{Machine Learning}
\label{sssec_tools_ml}

The top 5 machine learning tools used across all studies are listed below:
\begin{itemize}
	\item Weka\footnote{https://www.cs.waikato.ac.nz/ml/weka/}: a collection of machine learning algorithms for data mining tasks, including tools for data preparation, classification, regression, clustering, association rules mining and visualisation;
	\item scikit-learn\footnote{https://scikit-learn.org/}: consists of a set of tools for data mining and analysis, such as classification, regression, clustering, dimensionality reduction, model selection and pre-processing;
	\item LIBSVM\footnote{https://www.csie.ntu.edu.tw/\~cjlin/libsvm/}: an integrated software for support vector classification, regression, distribution estimation and multi-class classification;
	\item LIBLINEAR\footnote{https://www.csie.ntu.edu.tw/\~cjlin/liblinear/}: a linear classifier for data with millions of instances and features;  
	\item SVM-Light\footnote{http://svmlight.joachims.org/}: is an implementation of SVMs for pattern recognition, classification, regression and ranking problems.
\end{itemize}

\subsubsection{Opinion Mining}
\label{sssec_tools_om}

Certain studies used opinion mining tools in their research to either conduct their main experiments or for comparison purposes to their proposed solution/s. The following are the top 3 opinion mining tools used:
\begin{itemize}
    \item SentiStrength\footnote{http://sentistrength.wlv.ac.uk/}: a sentiment analysis tool that is able to conduct binary (positive/negative), trinary (positive/neutral/negative), single-scale (-4 very negative to very positive +4), keyword-oriented and domain-oriented classifications;
	\item Sentiment140\footnote{http://www.sentiment140.com/}: a tool that allows you to discover the sentiment of a brand, product, or topic on Twitter;
	\item VADER (Valence Aware Dictionary and sEntiment Reasoner)\footnote{https://github.com/cjhutto/vaderSentiment}: a lexicon and rule-based sentiment analysis tool that is specifically focused on sentiments expressed in social media.
\end{itemize}

\subsubsection{Big Data}
\label{sssec_tools_big_data}

Several big data technologies were used by the analysed studies. The most popular ones are categorised in the list below:
\begin{enumerate}
	\item Relational storage
	\begin{enumerate}
		\item MySQL\footnote{https://www.mysql.com/}
		\item PostgreSQL\footnote{https://www.postgresql.org/}
		\item Amazon Relational Database Service (Amazon RDS)\footnote{https://aws.amazon.com/rds/}
		\item Microsoft SQL Server\footnote{https://www.microsoft.com/en-us/sql-server/sql-server-downloads}
	\end{enumerate}
	\item Non-relational storage
	\begin{enumerate}
		\item Document-based
		\begin{enumerate}
			\item MongoDB\footnote{https://www.mongodb.com/}
			\item Apache CouchDB\footnote{http://couchdb.apache.org/}
		\end{enumerate}
		\item Column-based
		\begin{enumerate}
			\item Apache HBase\footnote{https://hbase.apache.org/} 
		\end{enumerate}
	\end{enumerate}
	\item Resource Description Framework Triplestore 
	\item Distributed Processing
	\begin{enumerate}
		\item Apache Hadoop\footnote{https://hadoop.apache.org/} 
		\item Apache Spark\footnote{https://spark.apache.org/} 
		\item IBM InfoSphere Streams\footnote{https://www.ibm.com/developerworks/library/bd-streamsintro/index.html} 
		\item Apache AsterixDB\footnote{https://asterixdb.apache.org/} 
		\item Apache Storm\footnote{https://storm.apache.org/}
	\end{enumerate}
	\item Data Warehouse
	\begin{enumerate}
		\item Apache Hive\footnote{https://hive.apache.org/}
	\end{enumerate}
	\item Data Analytics
	\begin{enumerate}
		\item Databricks\footnote{https://databricks.com/}
	\end{enumerate}
\end{enumerate}

The MySQL relational database management system was the most technology used for storing structured social data, whereas MongoDB was mostly used for processing unstructured social data. On the other hand, the distributed processing technologies were used for processing large scale social real-time and/or historical data. In particular, Hadoop MapReduce was used for parallel processing of large volumes of structured, semi-structured and unstructured social datasets, that are stored in the Hadoop Distributed File System. Spark's ability to process both batch and streaming data was utilised in cases where velocity is more important than volume.

\subsection{Natural Language Processing Tasks}
\label{ssec_nlp}

This section presents information about other NLP tasks that were conducted to perform SOM.

\subsubsection{Overview}
\label{sssec_overview}

An element of NLP is performed in 283 studies, out of the 465 analysed, either for pre-processing (248 studies), feature extraction (Machine Learning) or one of the processing parts within their SOM solution. The most common and important NLP tasks range from Tokenisation, Segmentation and POS, to NER and Language Detection.

It is important to mention that the NLP tasks mentioned above together with Anaphora Resolution, Parsing, Sarcasm, and Sparsity, are some other challenges faced in the SOM domain \cite{Khan2014}. Moreover, online posts with complicated linguistic patterns are challenging to deal with \cite{Li2014}. 

However, the authors in \cite{Koto2015-2} showcase the importance and potential of NLP within this domain, where they investigated the pattern or word combination of tweets in subjectivity and polarity by considering their POS sequence. Results reveal that subjective tweets tend to have word combinations consisting of adverb and adjective, whereas objective tweets tend to have a word combination of nouns. Moreover, negative tweets tend to have a word combination of affirmation words which often appear as a negation word.  

\subsubsection{Pre-processing and negations}
\label{sssec_preproc_neg}
The majority (355 out of 465) of the studies performed some sort of pre-processing in their studies. Different methods and resources were used for such a process, such as NLP tasks (e.g., tokenisation, stemming, lemmatisation, NER),  and dictionaries for stop words, acronyms for slang words, and others (e.g., noslang.com, noswearing.com, Urban Dictionary, Internet lingo).  

Negation handling is one of the most challenging issues faced by SOM solutions. However, 117 studies cater for negations within their approach, Several different methods are used, such as negation replacement, negation transformation, negation dictionaries, textual features based on negation words and negation models.  

\subsubsection{Emoticons/Emojis}
\label{sssec_emoticons}

Social media can be seen as a sub-language that uses emoticons/emojis mixed with text to show emotions \cite{Min2013}. Emoticons/emojis are commonly used in tweets irrespective of the language, therefore are sometimes considered as being domain and language independent \cite{Khan2014}, thus useful for multilingual SOM \cite{Cui2011}.

Even though some researchers remove emoticons/emojis as part of their pre-processing stage (depending on what the authors want to achieve), many others have utilised the respective emotional meaning within their SOM process. This has led to emoticons/emojis in playing a very important role within 205 solutions of the analysed studies especially when the focus is on emotion recognition. 

Results obtained from the emoticon networks model in \cite{ZhangL2013} show that emoticons can help in performing sentiment analysis. This is further supported by \cite{Jiang2015} who found that emoticons are a pure carrier of sentiment.  This is further supported by the results obtained by the emoticon polarity-aware method in \cite{LiD2018} which show that emoticons can significantly improve the precision for identifying the sentiment polarity. In the case of hybrid (lexicon and machine learning) approaches, emoticon-aided lexicon expansion improve the performance of lexicon-based classifiers \cite{Zhou2014}. From an emotion classification perspective, Porshnev et al. \cite{Porshnev2014b} analysed users' emoticons on Twitter to improve the accuracy of predictions for the Dow Jones Industrial Average and S\&P 500 stock market indices. Other researchers \cite{Cvijikj2011} were interested in analysing how people express emotions, displayed via adjectives or usage of internet slang i.e., emoticons, interjections and intentional misspelling. 

Several emoticon lists were used in these studies, with the Wikipedia and DataGenetics\footnote{http://www.datagenetics.com/blog/october52012/index.html} ones commonly used. Moreover, emoticon dictionaries, such as \cite{Agarwal2011, Aisopos2012, Becker2013}, consisting of emoticons and their corresponding polarity class were also used in certain studies.

\subsubsection{Word embeddings}
\label{sssec_word_embeddings}

Word embeddings, a type of word representation which allows words with a similar meaning to have a similar representation, were used by several studies \cite{Severyn2015, Jiang2015, Castellucci2015a, Castellucci2015b, Cai2015, Gao2015, ChenP2015, Stojanovski2015, GaoFei2016, Zhao2016, Rexha2016, Hao2017, Kitaoka2017, Arslan2018, Baccouche2018, ChenYuxiao2018, Ghosal2018, Hanafy2018, Jianqiang2018, Stojanovski2018, SunXiaoZhang2018, Wan2018, Yan2018} adopting a learning-based (Machine Learning, Deep Learning and Statistical) or hybrid approach. These studies used word embedding algorithms, such as word2vec, fastText\footnote{https://fasttext.cc/}, and/or GloVe\footnote{https://nlp.stanford.edu/projects/glove/}. Such a form of learned representation for text is capable of capturing the context of words within a piece of text, syntactic patterns, semantic similarity and relation with other words, amongst other word representations. Therefore, word embeddings are used for different NLP problems, with SOM being one of them.

\subsubsection{Aspect-based Social Opinion Mining}
\label{sssec_aspect_sa}

Sentence-level SOM approaches tend to fail in discovering an opinion dimension, such as sentiment polarity about a particular entity and/or its aspects \cite{Cambria2013}. Therefore, an aspect-level (also referred to as feature/topic-based) \cite{Hu2004} approach --where an opinion is made up of targets and their associated opinion dimension (e.g., sentiment polarity)-- has been used in some studies to overcome such issues. Certain NLP tasks, such as a parsing, POS tagger, and NER, are usually required to extract the entities or aspects from the respective social data.

From all the studies analysed, 39 performed aspect-based SOM, with 37 \cite{Bansal2018, Dragoni2018, Gandhe2018, Ghiassi2018, Kao2018, Katz2018, LiuQuanchao2018, Rathan2018, WangHongwei2018, Zainuddin2018, AbdullahNSD2017, Dambhare2017, Hagge2017, Ray2017, Rout2017, Tong2017, Vo2017, Zhou2017, Zimbra2016, Zainuddin2016a, Zainuddin2016b, Kokkinogenis2015, Lima2015, Hridoy2015, Castellucci2015a, Averchenkov2015, Tan2014, Lau2014, DelBosque2014, Varshney2014, Unankard2014, Lek2013, Wang2013, Min2013, Kontopoulos2013, Jiang2011, Prabowo2009} focusing on aspect-based sentiment analysis, 1 \cite{Aoudi2018} on aspect-based sentiment and emotion analysis and 1 \cite{Weichselbraun2017} on aspect-based affect analysis. 

In particular, the Twitter aspect-based sentiment classification process in \cite{Lek2013} consists of the following main steps: aspect-sentiment extraction, aspect ranking and selection, and aspect classification, whereas Lau et al. \cite{Lau2014} use NER to parse product names to determine their polarity. The aspect-based sentiment analysis approach in \cite{Hagge2017} leveraged POS tagging and dependency parsing. Moreover, \cite{Zainuddin2016a} proposed a hybrid approach to analyse aspect-based sentiment of tweets. As the authors claim, it is more important to identify the opinions of tweets rather than finding the overall polarity which might not be useful to organisations. In \cite{Zainuddin2018}, the same authors used association rule mining augmented with a heuristic combination of POS patterns to find single and multi-word explicit and implicit aspects. Results in \cite{Jiang2011} show that classifiers incorporating target-dependent features significantly outperform target-independent ones. In contrast to the studies discussed, Weichselbraun et al. \cite{Weichselbraun2017} introduced an aspect-based analysis approach that integrates affective (includes sentiment polarity and emotions) and factual knowledge extraction to capture opinions related to certain aspects of brands and companies. The social data analysed is classified in terms of sentiment polarity and emotions, aligned with the ``Hourglass of Emotions" \cite{Susanto2020}.

In terms of techniques, the majority of the aspect-based studies used a hybrid approach, where only 5 studies used deep learning for such a task. In particular, the study by Averchenkov et al. \cite{Averchenkov2015} used a deep learning approach based on RNNs for aspect-based sentiment analysis. A comparative review of deep learning for aspect-based sentiment analysis published by Do et al. \cite{Do2019} discusses current research in this domain. It focuses on deep learning approaches, such as CNN, LSTM and GRU, for extracting both syntactic and semantic features of text without the need for in-depth requirements for feature engineering as required by classical NLP. For future research directions on aspect-based SOM, refer to Section \ref{ssec_conc}.

\section{Dimensions of Social Opinion Mining}
\label{sec_social_opinion_mining-forms}

\subsection{Context}
\label{ssec_som_context}

An opinion describes a viewpoint or statement about a subjective matter. In many research problems, authors assume that an opinion is more specific and of a simpler definition. For example, sentiment analysis is considered to be a type of opinion mining even though it is only focused on extracting the sentimental score from a given text. Social data contains a wealth of signals to mine where opinions can be extracted over time. Different types of opinions require different modes of analysis \cite{Agrawal2014}.
This leads to opinions being multi-dimensional semantic artefacts. In fact, Troussas et al. specify that ``emotions and polarities are mutually influenced by each other, conditioning opinion intensities and emotional strengths". Moreover, multiple studies applied different approaches, where the authors in \cite{Bravo2013} showed that a composition of polarity, emotion and strength features, achieve significant improvements over single approaches, whereas \cite{Koto2015-1} focused on finding the correlation between emotion --which can be differentiated by facial expression, voice intonation and also words-- and sentiment in social media. Similar in nature, the authors in \cite{Buscaldi2015} found out that finer-grained negative tweets potentially help in differing between negative feelings, e.g., fear (emotion). 

Furthermore, mood, emotions and decision making are closely connected \cite{Porshnev2014a}. Research on multi-dimensional sentiment analysis shows that human mood is very rich in social media, where a piece of text may contain multiple moods, such as calm and agreement \cite{Huang2015}. On the other hand, there are studies showing that one mood alone is already highly influential in encouraging people to rummage through Twitter feeds for predictive information. For example in \cite{Weiss2015}, ``calmness" was highly correlated with stock market movement. 
Different dimenions of opinions are also able to effect different entities, such as events. Results in \cite{Zhang2012} show a strong correlation between emergent events and public moods. In such cases, new events can be identified by monitoring emotional vectors in microblogs. Moreover, work in \cite{Thelwall2011} assessed if popular events are correlated with sentiment strength as it increases, which is likely the case.

All of the above motivates us to pursue further research and development on the identification of different opinion dimensions that are present within social data, such as microblogs, published across heterogeneous social media platforms. A more fine-grained opinion representation and classification of this social data shall lead to a better understanding of the messages conveyed, thus potentially influencing multiple application areas. Section \ref{sec_social_opinion_mining-areas} lists the application areas of the analysed studies.

\subsection{Different Dimensions of Social Opinions identified in the Review Analysis}
\label{ssec_som_review_analysis}

The analysed studies focused on different opinion dimensions, namely: objectivity/subjectivity, sentiment polarity, emotion, affect, irony, sarcasm and mood. These were conducted on different levels, such as, document-level, sentence-level, and/or feature/aspect-based, depending on the study. Same as for the techniques presented in Section \ref{ssec_techniques_approaches}, 465 studies were evaluated. The majority focused on one social opinion dimension, with 60 studies focusing on more than one; 58 on two dimensions, 1 on three dimensions, and 1 on four dimensions. In this regard, Table \ref{table:studies-so-dimensions} lists the different dimensions and respective studies. Most of the studies focused on sentiment analysis, specifically polarity classification. 

\begin{comment}
58 on two dimensions, namely 
\textit{subjectivity and sentiment} \cite{Jiang2011, Blenn2012, Bravo2013, Zhu2013, Wang2013, Cui2013, Li2013, Rui2013, Bravo2014, Tan2014, Garg2014, Abdul2014, Samoylov2014, Koto2015-1, Koto2015-2, Koto2015-3, Feng2015, Mansour2015, Wu2016c, Zainuddin2016b, Er2016, AbdullahNSD2017, Hao2017, Ahuja2017, Sahni2017, Moh2017, Dritsas2018, Gandhe2018, Nausheen2018},
\textit{sentiment and emotion} \cite{Cvijikj2011, Orellana2013, Sheth2014, Yuan2015, Orellana2015, Gallegos2016, Qaisi2016, Shukri2015, Munezero2015, Barapatre2016, Karyotis2017, Bouazizi2017, Radhika2017, AbdullahMalak2017-1, Zhang2017, Singh2018, Aoudi2018, PaiRajesh2018, Ghosal2018, Rout2018, DosSantos2018, Stojanovski2018}, 
\textit{sentiment and mood} \cite{Bollen2011},
\textit{sentiment and irony} \cite{Reyes2013},
\textit{sentiment and sarcasm} \cite{Unankard2014},
\textit{emotion and anger} \cite{Delcea2014, Cotfas2015},
\textit{irony and sarcasm} \cite{Fersini2015} and
\textit{sentiment and affect} \cite{Weichselbraun2017}; 
1 \cite{Jiang2015} on three dimensions, namely \textit{subjectivity, sentiment and emotion}, and
1 \cite{Bosco2013} on four dimensions, namely \textit{subjectivity, sentiment, emotion and irony}. 
\end{comment}

\begin{table} [h!tp]
	%\scriptsize
	\begin{center}
    	%\begin{tabular}{ | p{1.5cm} | p{1.5cm} | p{7.25cm} |}
    	\begin{tabular}{ | p{4.5cm} | p{6.5cm} |}
    	\hline
     		\textbf{Dimensions}  				& \textbf{Studies} \\ \hline
     	    subjectivity and sentiment polarity	& \cite{Jiang2011, Blenn2012, Bravo2013, Zhu2013, Wang2013, Cui2013, Li2013, Rui2013, Bravo2014, Tan2014, Garg2014, Abdul2014, Samoylov2014, Koto2015-1, Koto2015-2, Koto2015-3, Feng2015, Mansour2015, Wu2016c, Zainuddin2016b, Er2016, AbdullahNSD2017, Hao2017, Ahuja2017, Sahni2017, Moh2017, Dritsas2018, Gandhe2018, Nausheen2018} \\ \hline
     	    sentiment polarity and emotion		&  \cite{Cvijikj2011, Orellana2013, Sheth2014, Yuan2015, Orellana2015, Gallegos2016, Qaisi2016, Shukri2015, Munezero2015, Barapatre2016, Karyotis2017, Bouazizi2017, Radhika2017, AbdullahMalak2017-1, Zhang2017, Singh2018, Aoudi2018, PaiRajesh2018, Ghosal2018, Rout2018, DosSantos2018, Stojanovski2018}\\ \hline
		  	sentiment polarity and mood 		& \cite{Bollen2011} \\ \hline
            sentiment polarity and irony 		& \cite{Reyes2013} \\ \hline
			sentiment polarity and sarcasm      & \cite{Unankard2014} \\ \hline
			sentiment polarity and affect		& \cite{Weichselbraun2017} \\ \hline
			emotion and anger					& \cite{Delcea2014, Cotfas2015} \\ \hline
			irony and sarcasm					& \cite{Fersini2015} \\ \hline
     	    subjectivity, sentiment polarity and emotion	& \cite{Jiang2015} \\ \hline
     	    subjectivity, sentiment polarity, emotion and irony	& \cite{Bosco2013} \\ \hline  	                    
    	\end{tabular}
	\end{center}
	\caption{Studies focusing on two or more social opinion dimensions}
	\label{table:studies-so-dimensions}
\end{table}

The following sections present the different tasks conducted for each form of opinion mentioned above\footnote{Note that some level categories are dependant on the domain}. 

\subsubsection{Subjectivity}

Subjectivity determines whether a sentence expresses an opinion --in terms of personal feelings or beliefs-- or not, in which case a sentence expresses objectivity. Objectivity refers to sentences that express some factual information about the world \cite{Liu2010}.

\begin{enumerate}
\item subjectivity classification: 2-level
	\begin{enumerate}
	\item objective/subjective
	\item neutral/subjective
	\item opinionated/not opinionated
	\end{enumerate}
\item subjectivity classification: 3-level
	\begin{enumerate}
	\item objective/positive/negative
	\item objective/subjective/subjective\&objective 
	\end{enumerate}
\item subjectivity score
	\begin{enumerate}
	\item objective/subjective ranging from 0 (low) to 1 (high)
	\end{enumerate}
\end{enumerate}

In this domain, objective statements are usually classified as being neutral (in terms of polarity), whereas subjective statements are non-neutral. In the latter cases, sentiment analysis is performed to determine the polarity classification (more information on this below). However, it is important to clarify that neutrality and objectivity are not the same. Neutrality refers to situations whereby a balanced view is taken, whereas objectivity refers to factual based i.e., true statements/facts that are quantifiable and measurable.  

\subsubsection{Sentiment}

Sentiment determines the polarity (positive/negative/neutral) and strength/intensity (through a numeric rating score e.g., 1 to 5 stars, or level of depth e.g., low/high/medium) of an expressed opinion \cite{Liu2010}. 

\begin{enumerate}
\item polarity classification: 2-level
	\begin{enumerate}
	\item positive/negative
	\item for/against
	\item pro/against
	\item positive/not positive (neutral or negative)
	\end{enumerate}
\item polarity classification: 3-level
	\begin{enumerate}
	\item positive/negative/neutral
	\item positive/negative/null
	\item positive/negative/contradiction
	\item positive/negative/objective
	\item positive/negative/other (neutral, irrelevant)
	\item positive/negative/humorous
	\item positive/negative/irrelevant
	\item positive/negative/uncertain
	\item beneficial (positive)/harmful (negative)/neutral
	\item personal negative/personal non-negative/non-personal i.e. news
	\item bullish/bearish/neutral
	\end{enumerate}
\item polarity classification: 4-level
	\begin{enumerate}
	\item positive/not so positive/not so negative/negative
	\item positive/negative/neutral/irrelevant
	\item positive/negative/neutral/undefined
	\item positive/negative/neutral/none
	\item positive/negative/neutral/meaningless
	\item positive/negative/neutral/not related to target topic
	\item positive/negative/neutral/unsure
	\item positive/negative/neutral/uninformative
	\item subjective/positive/negative/ironic (subjectivity and irony classification is also considered)
	\item positive/negative/mixed/none
	\item for/against/mixed/neutral
	\item positive/negative/neutral/ideology/sarcastic
	\end{enumerate}
\item polarity classification: 5-level
	\begin{enumerate}
	\item highly positive/positive/neutral/negative/highly negative
	\item strong positive/positive/neutral/negative/strong negative
	\item strongly positive/mildly positive/neutral/mildly negative/strongly negative 
	\item strongly positive/slightly positive/neutral/slightly negative/strongly negative 
	\item very positive/positive/neutral/negative/very negative
	\item positive/somewhat positive/neutral/somewhat negative/negative
	\item most positive/positive/neutral/negative/most negative
	\item extremely positive/positive/neutral/negative/extremely negative
	\item positive/negative/ironic/positive and negative/objective (subjectivity and irony classification is also considered)
	\item excellent/good/neutral/bad/worst
	\item worst/bad/neutral/decent/wonderful
	\end{enumerate}
\item polarity classification: 6-level
	\begin{enumerate}	
	\item strong positive/steady positive/week positive/week negative/steady negative/strong negative
	\end{enumerate}
\item polarity classification: 8-level
	\begin{enumerate}
	\item partially positive/mildly positive/positive/extremely positive/partially negative/mildly negative/negative/extremely negative 
	\item ProCon/AntiCon/ProLab/AntiLab/ProLib/AntiLib/Unknown/Irrelevant (levels are oriented towards the political domain)
	\end{enumerate}
\item polarity classification: 12-level
	\begin{enumerate}
	\item future orientation/past orientation/positive emotions/negative emotions/sadness/anxiety/anger/
	tentativeness/certainty/work/achievement/money
	\end{enumerate}
\item polarity score
	\begin{enumerate}
	\item negative ranging from 0-0.5 and positive ranging from 0.5-1
	\item negative/neutral/positive ranging from 0 (low) to 0.45 (high)
	\item negative/positive ranging from -1 (low) to 1 (high) 
	\item negative/positive ranging from -1.5 (low) to 1.5 (high)
	\item negative/positive ranging from -2 (low) to 2 (high)
	\item negative/positive ranging from 1 (low) to 5 (high)
	\item negative ranging from -1 (low) to -5 (high) and positive ranging from 1 (low) to 5 (high)
	\item strongly negative to strongly positive ranging from -2 (low) to 2 (high)
	\item normalised values from -100 to 100 
	\item weighted average of polarity scores of the sentiment aspects/topic segments 
	\item score for every aspect/feature of the subject
	\item  score per aspect by calculating the distance between the aspect and sentiment word
	\item total classification probability close to 1
	\end{enumerate}
\item polarity strength
	\begin{enumerate}
	\item -5 (very negative) to 5 (very positive)
	\item 1 (no sentiment) to 5 (very strong positive/negative sentiment)
	\item low (0) to high (5)
	\item -4 (most negative) to 4 (most positive)
	\item weak/strong (relative strength)
	\item Euclidean distance of positive and negative dimensions 
	\end{enumerate}
\item polarity intensity
	\begin{enumerate}
	\item normal/excited/passionate
	\item no emotion/a bit/normal/very/extremely
	\item -3 (negative) to 3 (positive)
	\end{enumerate}
\item sentiment assignment
	\begin{enumerate}
	\item total sentiment is the sum of sentiment of all words divided by total number of words (high to low)
	\item average mean sentiment score
	\item sentiment index based on the distribution of positive and negative online posts \cite{Oh2017}
	\item sum of inverse distance weighted sentiment values (+1, -1) of words in textual interactions 
	\item sentiment for a term is computed as [min, max] of all the positive and negative polarities
	\item average score of associated messages in a time range and overall sentiment trend encoded by colours
	\end{enumerate}
\item other
	\begin{enumerate}
	\item cluster heads from sentimental content
	\item sentiment change detection
	\end{enumerate}
\end{enumerate}

In some studies \cite{Sandoval2018, Bouazizi2017, Chou2017, Karyotis2017, Furini2016, Gambino2016, Jiang2015, Yuan2015}, the sentiment polarity was derived from the emotion classification, such as, joy/love/surprise translated to positive, and anger/sadness/fear translated to negative.

\subsubsection{Emotion}

Emotion refers to a person's subjective feelings and thoughts, such as love, joy, surprise, anger, sadness and fear \cite{Liu2010}. 

\begin{enumerate}
\item emotion classification: 2-level
	\begin{enumerate}
	\item happy/sad
	\item emotional/non-emotional
	\end{enumerate}
\item emotion classification: 3-level
	\begin{enumerate}
	\item happy/sad/neutral
	\end{enumerate}
\item emotion classification: 4-level
	\begin{enumerate}
	\item happy/sad/angry/surprise
	\item happy/sad/angry/calm
	\item joy/sadness/anger/fear
	\end{enumerate} 
\item emotion classification: 5-level
	\begin{enumerate}
	\item nervous/sympathetic/ashamed/worried/angry
	\item happy/sad/angry/laughter/scared
	\item happy/surprised/sad/angry/neutral
	\end{enumerate}
\item emotion classification: 6-level
	\begin{enumerate}
	\item joy/sadness/surprise/anger/fear/disgust
	\item joy/sadness/fear/anger/disgust/unknown
	\item joy/sadness/fear/anger/surprise/unknown
	\item fear/anger/disgust/happiness/sadness/surprise
	\end{enumerate}
\item emotion classification: 7-level
	\begin{enumerate}
	\item anger/disgust/fear/happy/neutral/sarcastic/surprise
	\item pleasure/wondering/confirmation/excitement/laughter/tasty/surprise (emotions based on interjections \cite{Cvijikj2011}) 
	\item love-heart/quality/happiness-smile/sadness/amused/anger/thumbs up (emotions based on sentiment carrying words and/or emoticons \cite{Walha2016}) 
	\item joy/surprise/sadness/fear/anger/disgust/unknown
	\item like/happiness/sadness/disgust/anger/surprise/fear
	\item joy/love/anger/sadness/fear/dislike/surprise
	\item anger/joy/love/fear/surprise/sadness/disgust
	\item joy/sadness/anger/love/fear/thankfulness/surprise
	\item happiness/sadness/anger/disgust/fear/surprise/neutral
	\item joy/surprise/fear/sadness/disgust/anger/neutral
	\item love/happiness/fun/neutral/hate/sadness/anger
	\item happiness/goodness/anger/sorrow/fear/evil/amazement
	\end{enumerate}
\item emotion classification: 8-level
	\begin{enumerate}
	\item anger/embarrassment/empathy/fear/pride/relief/sadness/other
	\item flow/excitement/calm/boredom/stress/confusion/frustration/neutral
	\item joy/sadness/fear/anger/anticipation/surprise/disgust/trust
	\item anger/anxiety/expect/hate/joy/love/sorrow/surprise
	\item happy/loving/calm/energetic/fearful/angry/tired/sad
	\item anger/sadness/love/fear/disgust/shame/joy/surprise
	\end{enumerate}
\item emotion classification: 9-level
	\begin{enumerate}
	\item surprise/affection/anger/bravery/disgust/fear/happiness/neutral/sadness
	\end{enumerate}
\item emotion classification: 11-level
	\begin{enumerate}
	\item neutral/relax/docile/surprise/joy/contempt/hate/fear/sad/anxiety/anger
	\item joy/excitement/wink/happiness/love/playfulness/surprise/scepticism/support 
	/sadness/annoyance (emotions based on emoticons \cite{Cvijikj2011})
	\end{enumerate}
\item emotion classification: 22-level
	\begin{enumerate}
	\item hope/fear/joy/distress/pride/shame/admiration/reproach/ 
	 
	 linking/disliking/gratification/remorse/gratitude/anger/ satisfaction/

	 fears-confirmed/relief/disappointment/happy-for/resentment/gloating/pity (emotions based on an Emotion-Cause-OCC model that describe the eliciting conditions of emotions \cite{Gao2015})
	\end{enumerate}
\item emotion - anger classification: 7-level
	\begin{enumerate}
	\item frustration/sulking/fury/hostility/indignation/envy/annoyance
	\end{enumerate}
\item emotion score
	\begin{enumerate}
	\item valence/arousal/dominance ranging from 1 (low) to 9 (high)
	\item prediction/valence/arousal/outcome from 0 (low) to 100 (high)
	\end{enumerate}
\item emotion intensity
	\begin{enumerate}
	\item 0 (minimum) to 1 (maximum)	
	\item 0 (minimum) to 9 (maximum)
	\item high/medium/low
	\end{enumerate}
\item emotion - happiness measurement
	\begin{enumerate}
	\item average happiness score
	\end{enumerate}
\end{enumerate}

A study \cite{Munezero2015} mapped the observed emotions into two broad categories of enduring sentiments: `like' and `dislike'. The former includes emotions that have a positive evaluation of the object, i.e., joy, trust and anticipation, and the latter includes emotions that have a negative evaluation of the object, i.e., anger, fear, disgust, and sadness.

It is important to note that some of the emotion categories listed above are based on published  theories of emotion, with the most popular ones being Paul Ekman's six basic emotions (anger, disgust, fear, happiness, sadness and surprise) \cite{Ekman1992},
and Plutchik's eight primary emotions (anger, fear, sadness, disgust, surprise, anticipation, trust, and joy) \cite{Plutchik1980}.
Moreover, other studies have used emotion categories that are influenced from emotional state/psychological models, such as the Pleasure Arousal Dominance \cite{Mehrabian1996} and the Ortony, Clore and Collins (commonly referred to as OCC) \cite{Ortony1988}.

Several studies \cite{Xu2012, Furini2016, Walha2016, Hubert2018} that targeted emotion classification incorrectly referred to such a task as sentiment analysis. Even though emotions and sentiment are highly related, the former are seen as enablers to the latter, i.e., an emotion/set of emotions affect the sentiment.

\subsubsection{Affect}

Affect refers to a set of observable manifestations of a subjectively experienced emotion. The basic tasks of affective computing are emotion recognition and polarity detection \cite{Cambria2016}.

\begin{enumerate}
\item affect classification: 4-level
	\begin{enumerate}	
	\item aptitude/attention/pleasantness/sensitivity (based on the ``Hourglass of Emotions", which was inspired by Plutchik's studies on human emotions)
	\end{enumerate}
\end{enumerate}

When using the affective model mentioned above, sentiment is based on the four independent dimensions mentioned, namely Pleasantness, Attention, Sensitivity, and Aptitude. The different levels of activation of these dimensions constitute the total emotional state of the mind \cite{Hussain2018}. The semi-supervised learning model proposed by Hussain and Cambria \cite{Hussain2018} based on the merged use of multi-dimensional scaling by means of random projections and biased SVM, has been exploited for the inference of semantics and sentics (conceptual and affective information) that are linked with concepts in a multi-dimensional vector space, in accordance with this affective model. This is used to carry out sentiment polarity detection and emotion recognition in cases when there is a lack of labelled common-sense data.

\subsubsection{Irony}

Irony is usually used to convey, the opposite meaning of the actual things you say, but its purpose is not intended to hurt the other person.

\begin{enumerate}
\item irony classification: 2-level
	\begin{enumerate}
	\item ironic/non-ironic
	\end{enumerate}
\end{enumerate}

\subsubsection{Sarcasm}

Sarcasm holds the ``characteristic" of meaning the opposite of what you say, but unlike irony, it is used to hurt the other person.

\begin{enumerate}
\item sarcasm classification: 2-level
	\begin{enumerate}
	\item sarcastic/non-sarcastic
	\item yes/no
	\end{enumerate}
\end{enumerate}

\subsubsection{Mood}

Mood refers to a conscious state of mind or predominant emotional state of person or atmosphere of groups, people or places, at any point in time.

\begin{enumerate}
\item mood classification: 6-level
	\begin{enumerate}	
	\item composed-anxious/agreeable-hostile/elated-depressed/confident-unsure
	      /energetic-tired/clearheaded-confused (based on the profile of mood states (POMS) Bipolar questionnaire \cite{McNair1971}
	 which is designed by psychologists to assess human mood states)
	\item calm/alert/sure/vital/kind/happy (based on GPOMS \cite{Bollen2011} which expands the POMS Bipolar questionnaire to capture a wider variety of naturally occurring mood terms in tweets) 
	\end{enumerate}
\item mood classification: 8-level
	\begin{enumerate}
	\item happy/loving/calm/energetic/fearful/angry/tired/sad
	\end{enumerate}
\end{enumerate}

\subsubsection{Aggressiveness}

The authors in \cite{DelBosque2014} assume that aggressive text detection is a sub-task of sentiment analysis, which is closely related to document polarity detection. Their reasoning is that aggressive text can be seen as intrinsically negative.

\begin{enumerate}
\item Aggressiveness detection
	\begin{enumerate}
	\item aggressiveness score ranging from 0 (no aggression) to 10 (strong aggression)
	\end{enumerate}
\end{enumerate}

\subsubsection{Other}

\begin{enumerate}
\item Opinion retrieval
	\begin{enumerate}
	\item opinion score from 0 (minimum) to 5 (maximum)
	\end{enumerate}
\end{enumerate}

\subsection{Impact of Sarcasm and Irony on Social Opinions}
\label{ssec_impact_sar_iro}

Sarcasm and irony are often confused and/or misused. This leads to their classification in being very difficult even for humans \cite{Unankard2014, Buscaldi2015}, with most users holding negative views on such messages \cite{Unankard2014}. The study by Buscaldi and Hernandez-Farias \cite{Buscaldi2015} is a relevant example, whereby a large number of false positives were identified in the tweets classified as ironic. Moreover, such tasks are also very time consuming and labour intensive particularly with the rapid growth in volume of online social data. Therefore, not many studies focused and/or catered for sarcasm and/or irony detection. 

\subsubsection{Challenges}

The majority of the reviewed proposed approaches are not equipped to cater for traditional limitations, such as negation effects or ironic phenomena in text \cite{Castellucci2015a}. Such opinion mining tasks face several challenges, with the main ones being:
\begin{itemize}
	\item Different languages and cultures result in various ways of how an opinion is expressed on certain social media platforms. For example, Sina Weibo users prefer to use irony when expressing negative polarity \cite{WangZhitao2014}. Future research is required for the development of cross-lingual/multilingual NLP tools that are able to identify irony and sarcasm \cite{YanG2014}.  
	\item Presence of sarcasm and irony in social data, such as tweets, may affect the feature values of certain machine learning algorithms. Therefore, further advancement is required in the techniques used for handling sarcastic and ironic tweets \cite{Pandey2017}. The work in \cite{Sarsam2020} addresses the main challenges faced for sarcasm detection in Twitter and the machine learning algorithms that can be used in this regard.	
	\item Classifying/rating a given sentence's sentiment is very difficult and ambiguous, since people often use negative words to be humorous or sarcastic. 
	\item Sarcasm and/or irony annotation is very hard for humans and thus it should be presented to multiple persons for accuracy purposes. This makes it very challenging to collect large datasets that can be used for supervised learning, with the only possible way being to hire people to carry out such annotations \cite{DAsaro2017}. Moreover, the differentiation between sarcasm and irony by human annotators result in a lack of available datasets and datasets with enough examples of ironic and/or sarcastic annotations. Such datasets are usually needed for ``data hungry" computational learning methods \cite{Sykora2020}. 	
\end{itemize}

\subsubsection{Observations}

Table \ref{table:studies-sa-ir} lists the studies within the review analysis that focused on sarcasm and/or irony. These account for only 18 out of 465 reviewed papers. One can clearly note the research gap that exists within these research areas. 

%\vspace*{-3mm}

\begin{table} [h!tp]
	\small
	\begin{center}
    	%\begin{tabular}{ | p{1.5cm} | p{1.5cm} | p{7.25cm} |}
    	\begin{tabular}{ | c | c | c |}
    	\hline
     	             \textbf{Sarcasm} & \textbf{Irony}  & \textbf{Studies} \\ \hline
     	             \cmark  &        & \cite{Baccouche2018, Bouazizi2018, Ghiassi2018, AbdullahNSD2017, Bouazizi2017, Caschera2016, Tan2014, Unankard2014, Mejova2013, Bakliwal2013, Mejova2012, WangH2012} \\ \hline
     	                     & \cmark & \cite{Buscaldi2015, Hernandez2014, Bosco2013, Reyes2013}   \\ \hline
     	             \cmark  & \cmark & \cite{Fersini2015, Pandey2017} \\ \hline     	            
    	\end{tabular}
	\end{center}
	\caption{Studies adopting sarcasm and/or irony}
	\label{table:studies-sa-ir}
\end{table}

%\vspace*{-8mm}

The following is an overview of the studies' main results and observations: 
\begin{itemize}
	\item \cite{Bosco2013}: The authors found that irony is normally used together with a positive statement to express a negative statement, but seldomly the other way. Analysis shows that the Senti-TUT\footnote{http://www.di.unito.it/\~tutreeb/sentiTUT.html} corpus can be representative for a wide range of irony in phenomena from bitter sarcasm to genteel irony.  
	\item \cite{Reyes2013}: The study describes a number of textual features used to identify irony at a linguistic level. These are mostly applicable for short texts, such as tweets. The developed irony detection model is evaluated in terms of representativeness and relevance. Authors also mention that there are overlaps in occurrences of irony, satire, parody and sarcasm, with their main differentiators being tied to usage, tone and obviousness.
	\item \cite{Mejova2013}: A multi-stage data-driven political sentiment classifier is proposed in this study. The authors found out ``that a humorous tweet is 76.7\% likely to also be sarcastic", whereas ``sarcastic tweets are only 26.2\% likely to be humorous". Future work is required on the connection between sarcasm and humour. 
	\item \cite{Fersini2015}: Addresses the automatic detection of sarcasm and irony by introducing an ensemble approach based on Bayesian Model Averaging, that takes into account several classifiers according to their reliabilities and their marginal probability predictions. Results show that not all the features are equally able to characterise sarcasm and irony, whereby sarcasm is better characterised by POS tags, and ironic statements by pragmatic particles (such as emoticons and emphatic/onomatopoeic expressions, which represent those linguistic elements typically used in social media to convey a particular message).
	\item \cite{Jiang2015}: The authors' model classifies subjectivity, polarity and emotion in microblogs. Results show that emoticons are a pure carrier of sentiment, whereas sentiment words have more complex senses and contexts, such as negations and irony. 
	\item \cite{WangH2012}: Post-facto analysis of user-generated content, such as tweets, show that political tweets tend to be quite sarcastic. 
	\item \cite{Ghiassi2018}: Certain keywords or hash-tagged words (e.g., ``thanks", ``\#smh", `` \#not") that follow certain negative or positive sentiment markers in textual social data, usually indicate the presence of sarcasm.
\end{itemize}

\section{Application Areas of Social Opinion Mining}
\label{sec_social_opinion_mining-areas}

\begin{figure}[!htb]
	\centering    
    \includegraphics[width=\textwidth]{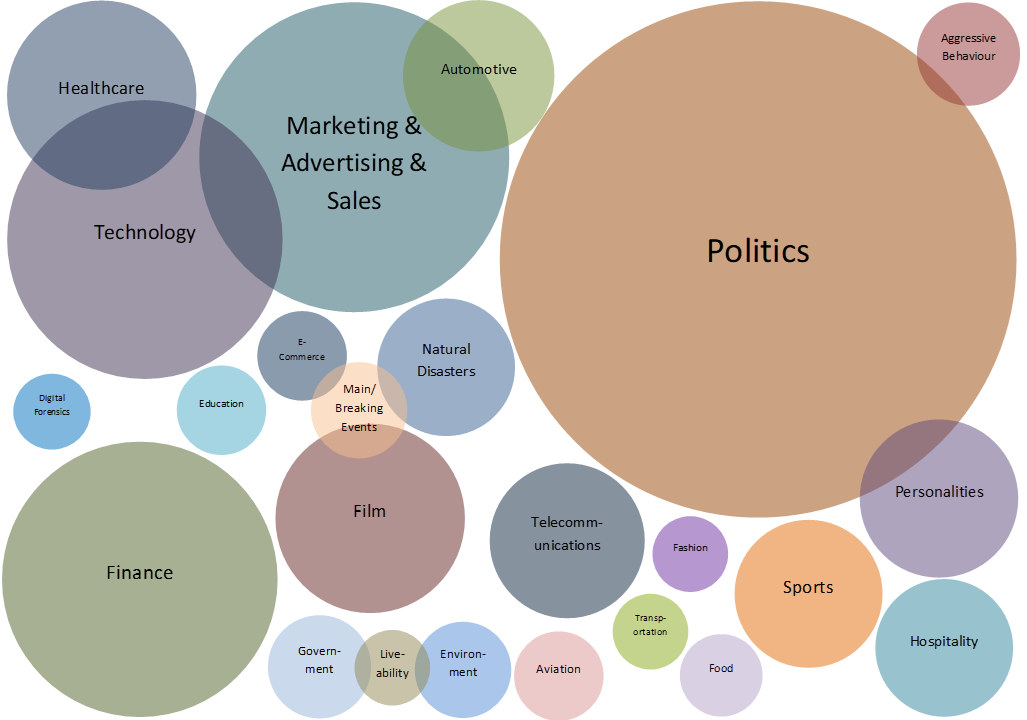}
	\caption{Application Areas}
	\label{fig:application-areas}
\end{figure}

Around half of the studies analysed focused their work on a particular real-world application area (or multiple), where Figure \ref{fig:application-areas} shows the ones applicable for this systematic review. Note that each circle represents an application area, where the size reflects the number of studies within the particular application area. The smallest circles represent a minimum of two studies that pertain to the respective application area, whereas the biggest circle reflects the most popular application area. Intersecting circles represent application areas that were identified as being related to each other based on the analysis conducted.

%\newpage

The \textbf{Politics} domain is the dominant application area with 45 studies applying SOM on different events, namely 
elections \cite{Elouardighi2017, Bansal2018, Nugroho2017, ChenYang2018, Nausheen2018, AbdullahMalak2017-1, Joyce2017, Soni2017, Salari2018, Fatyanosa2017, Juneja2017, Sandoval2018, Zhou2017, Le2017, Yuan2014, Ramteke2016, Smailovic2015, Burnap2016, Rill2014, Anjaria2014, Kuo2016, Batista2014, Mejova2013, Hoang2013, Goncalves2013, Unankard2014, WangH2012, Maynard2011, Bosco2013, Bakliwal2013, Tumasjan2010}, reforms, such as equality marriage \cite{Lai2015}, debates \cite{Tapia2014}, referendums \cite{Pavel2017, Fang2017}, political parties or politicians \cite{Ozer2017, Javed2014, Taddy2013}, and political events, such as terrorism, protests, uprisings and riots \cite{Sachdeva2018, Kamyab2018, Bouchlaghem2016, Mejova2012, deSouzaCarvalho2016, Sheth2014, Weiss2013}. 

In terms of \textbf{Marketing \& Advertising \& Sales}, 29 studies focused on brand/product management and/or awareness \cite{Giachanou2017, Ayoub2018, Ghiassi2018, LiYujiao2018, Ducange2017, Husnain2017, Teixeira2017, Halibas2018, Hu2017, AbdullahNSD2017, Zimbra2016, Cho2014, Esiyok2015, Dasgupta2015, Ghiassi2013, Mostafa2013, Min2013, Cvijikj2011, Li2013, Goncalves2013}, products/services in general \cite{Asghar2018, Kao2018, Walha2016, Polymerou2014, Li2013}, local marketing \cite{Costa2014} and online advertising \cite{Adibi2018, Dragoni2018, Lewenberg2015}. 

The \textbf{Technology} industry-oriented studies (23) focused on either: company perception \cite{Wan2018, Rout2018, Lek2013, Petz2013, Jiang2011}, products, such as mobile/smart phones \cite{Rathan2018, Ray2017, Geetha2018, GuptaI2017, Gandhe2018, Hridoy2015, Agrawal2014, Suresh2016, Mumu2014, Erdmann2014}, laptops \cite{Raja2016}, electronics \cite{Neethu2013} tablets \cite{Severyn2016, Jiang2011}, operating systems \cite{Huang2018}, cloud service providers \cite{Qaisi2016}, social media providers \cite{Arslan2017} and multiple technologies \cite{Vo2017}. 

All the 21 studies targeting the \textbf{Finance} domain applied SOM on demonitisation \cite{GuptaF2017}, currencies \cite{Pavel2017} and the stock market, for risk management \cite{Ishikawa2017} and predictive analytics \cite{Ghosal2018, Chen2018, Pineiro2018, Simoes2017, Tiwari2017, Sun2017, Coyne2017, Zhao2016, Attigeri2015, Weiss2015, Rao2014, Huang2015, Porshnev2013, Porshnev2014a, Porshnev2014b, Yu2013, Bollen2011, Vu2012}.

Thirteen studies applied SOM on the \textbf{Film} industry for recommendations \cite{Orellana2015, Song2015}, box office predictions \cite{Du2014, Rui2013} or from a general perspective \cite{Pavel2017, Sihwi2018, Permatasari2018, Orellana2013, YanG2014, Goncalves2013, Wang2013, Blenn2012, Chen2012}. Similarly, 13 studies focused on \textbf{Healthcare}, namely on epidemics/infectious diseases \cite{Hong2018, Lim2017, LuY2015, Goncalves2013}, drugs \cite{Moh2017, Peng2016, Wu2015}, hospitals \cite{Gupta2016}, vaccines \cite{Song2017}, public health, such as epidemics, clinical science and mental health \cite{Ji2015, Ji2016}, and in general, such as health-related tweets \cite{Baccouche2018} and health applications \cite{PaiRajesh2018}.

In terms of other industries, SOM was applied within the following:
\begin{itemize}
	\item \textbf{Telecommunications} (e.g., telephony, television) on particular service providers \cite{Ghiassi2018, Ranjan2018, Napitu2017, Fitri2018, Kumar2016, Varshney2014, Wunnasri2013, Tan2011, Trung2013} or complaints \cite{Souza2016}; 
	\item \textbf{Automotive} \cite{Vo2017, PaiPing2018, Fatyanosa2018, Weichselbraun2017, Shukri2015, Bifet2011, Reyes2013, Severyn2016, Erdmann2014};
	\item \textbf{Hospitality} for restaurant recommendations \cite{Vo2017, Yang2013} and hotel/resort perceptions \cite{Rout2017, Philander2016, Lu2016, ChenC2015, Molina2014};
	\item \textbf{Aviation} on specific airline services, e.g., customer relationship management \cite{Ghiassi2018, Mostafa2013b, Chen2016}, and air crashes \cite{Goncalves2013};
	\item \textbf{Food} either in general \cite{DosSantos2018, Liu2015} or on safety \cite{Sun2014};
	\item \textbf{Fashion} \cite{Mukkamala2014-1, Mukkamala2014-2}.	
\end{itemize}

In terms of domains, the studies focused on: 
\begin{itemize}
	\item \textbf{Sports} on football/soccer \cite{Stojanovski2018, Seron2015, Guerra2014}, American football \cite{Guerra2014, Brooks2014}, basketball \cite{Tan2011, Jiang2011}, cricket \cite{Ahuja2017} and Olympics \cite{Goncalves2013};
	\item \textbf{Government} for smart cities \cite{DAsaro2017, Anggoro2016, LiM2016} and e-Government \cite{Hubert2018, Rezk2018, Williamson2016};
	\item \textbf{Environment} for policy makers \cite{Sluban2015}, urban mobility \cite{Gallegos2016}, wind energy \cite{Politopoulou2013}, green initiatives \cite{Rai2018} and peatland fires \cite{Gandhe2018};
	\item \textbf{E-commerce} for product recommendations \cite{Xie2012, Lau2014}, crisis management \cite{Park2011}, decision making \cite{Davanzo2015} and policy making \cite{Omar2017};
	\item \textbf{Education} for e-learning \cite{Ortigosa2014, Karyotis2017} and on universities \cite{Abdelrazeq2016};
	\item \textbf{Transportation} for ride hailing services and logistics \cite{Anastasia2016} and traffic conditions \cite{Cao2018}.	
\end{itemize} 

Moreover, other studies applied SOM in the following areas:
\begin{itemize}
	\item \textbf{Personalities} \cite{Ali2018, Ghiassi2018, Arslan2017, Tasoulis2018, Goel2018, Poortvliet2018, WangHongwei2018, Jiang2011, Tan2011, Khan2014, Kranjc2013};	
	\item \textbf{Natural Disasters} on earthquakes \cite{Aoudi2018, Ragavi2014, Zhang2012, Thelwall2011}, flooding \cite{Buscaldi2015}, explosions \cite{Ouyang2017} and in general \cite{Sangameswar2017};
	\item \textbf{Aggressive Behaviour} in relation to crime \cite{Kitaoka2017, ChenX2015, Zainuddin2016a}, cyberbullying \cite{DelBosque2014}, bullying \cite{Xu2012} and violence and disorder \cite{Jurek2014};	
    \item \textbf{Main/Breaking Events} such as Black Friday \cite{Choi2013}, Oscars, TV shows, product launch, earthquake \cite{Thelwall2011}, accidents e.g., shootings \cite{Singh2018, Akcora2010} and in general \cite{Stojanovski2018};
	\item \textbf{Liveability} in terms of place design to supports local authorities, urban designers and city planners \cite{You2016}, and government services, such as welfare \cite{Flaes2016};	
    \item \textbf{Digital Forensics} \cite{Andriotis2014, Aboluwarin2016}.
\end{itemize}

Lastly, 19 further studies --not represented in Figure \ref{fig:application-areas}-- focused on the following application areas: 
Human Development \cite{Zafar2016}, 
Human Mobility \cite{Kokkinogenis2015}, 
Public Facilities \cite{Ramadhani2016}, 
Smart Cities \cite{Li2017}, 
Web Publishing \cite{Tian2015}, 
Sponsorships \cite{Kaushik2016}, 
Countries \cite{Khan2014}, 
Industry \cite{Trung2013}, 
Entertainment \cite{Trung2013}, 
Refugee/Migrant crisis \cite{Lee2017}, 
Tourism \cite{Michailidis2018}, 
Music \cite{Radhika2017}, 
Cryptocurrency \cite{Pant2018}, 
Economy \cite{GuptaF2017}, 
Social Issues \cite{Vora2017}, 
Law \cite{Gandhe2018}, 
Insurance/Social Security \cite{Zhang2017}, 
Geographic Information \cite{Stojanovski2018} 
and Social Interactions \cite{Vivanco2017}.

\section{Concluding Remarks}
\label{sec_conc_rem}

This section presents the latest research developments and advancements within the SOM research area (Section \ref{ssec_som_latest}) and presents the overall conclusions of this systematic review in terms of target audience and future research and development in (Section \ref{ssec_conc}).

\subsection{Latest Research of Social Opinion Mining}
\label{ssec_som_latest}

Given that this systematic review covers studies till 2018, some recent developments and advancements from 2019 till 2021 shall be discussed within this sub-section. This shows the fast research turnaround in SOM which has kept evolving at an incredibly fast rate, thus reiterating its validity and popularity as a research area.

The number of studies using Deep Learning approaches continued to increase (as reflected in Table \ref{table:om-approaches}), especially ones using certain deep learning techniques, such as CNNs, RNNs, LSTM, GRU and Deep Belief Networks \cite{Yadav2020, Wadawadagi2020}, and with the introduction of new techniques, such as Transfer Learning. This is supported by numerous studies \cite{Carvalho2021, Eke2020} who have noted that researchers are shifting from using traditional machine learning techniques to deep learning ones. Carvalho and Plastino \cite{Carvalho2021} focus on sentiment analysis on tweets, Xu et al. \cite{Xu2020} focus on emotion classification on tweets, Akhtar et al. \cite{Akhtar2020} focus on sentiment and emotion intensity, Cignarella et al. \cite{Cignarella2020} focus on irony detection of English, Spanish, French and Italian tweets, whereas Eke et al. \cite{Eke2020} focus on sarcasm detection with Twitter also being the social media platform mostly used in this research area. 

Transfer learning is a deep learning technique where a model is trained for one or more tasks (source tasks), which learnt knowledge is applied to a related second task (target task) \cite{Pan2009}. In particular, the Transformer model architecture introduced by Vaswani et al. \cite{Vaswani2017} in 2017, is based on attention mechanisms and is designed to handle sequential data like natural language for NLP tasks, such as sentiment analysis and text summarisation. This has coincided with the advancement of SOM for different opinion dimensions, such as sentiment polarity \cite{Nguyen2020, Naseem2020}, emotion \cite{Acheampong2021}, and irony \cite{Nguyen2020}, especially studies focused on adaptation to new domains and/or knowledge transfer from one language to another. The latter application is extremely reliable for cross-lingual adaptation where a labelled dataset is available in one language e.g., English, which is then applied to another language, such as low-resourced languages \cite{Ruder2017transferlearning}.

With respect to language, more SOM studies supporting languages other than the popular ones (such as English and Chinese) are on the rise. In \cite{Rani2019}, the authors discuss the growth of research work in the fields of sentiment and emotion analysis for Indian languages. Moreover, Buechel et al. \cite{Buechel2020} created emotion lexicons for 91 languages for sentiment and emotion analysis. Other recent studies have focused on languages, such as Urdu for sentiment analysis \cite{Mukhtar2019}, Maltese for sentiment and emotion analysis and sarcasm/irony detection \cite{Cortis2019}, Indonesian for sentiment analysis \cite{Koto2020}, Portuguese for sentiment and emotion analysis \cite{Pereira2021}, and Arabic for sentiment and emotion analysis \cite{Alhumoud2021}. Studies on code-switched languages is also on the increase, with \cite{Bansal2020} demonstrating how Hindi-English code-switching patterns from tweets can be used to improve sarcasm detection, and \cite{Appidi2020} analysing code-switched Kannada-English from tweets for emotion classification. 

In terms of modality, the visual modality is gaining more interest in the SOM research community. In \cite{Akhtar2019}, the authors propose a deep multi-task learning framework that carries out sentiment and emotion analysis from the textual, acoustic and visual frames of video data obtained from YouTube. On the other hand, Kumar and Garg \cite{Kumar2019} propose a multi-modal sentiment analysis model for Twitter, where the sentiment polarity and strength is extracted from tweets based on their text and images (typographic and/or infographic). 

More research has been published on aspect-based SOM, with \cite{Jiang2020} focused on sentiment polarity in both single-aspect and multi-aspect scenarios, whereas \cite{Hyun2020} focused on sentiment polarity in the automotive domain for the English and Korean languages. 

In terms of application areas, the ones identified in Section \ref{sec_social_opinion_mining-areas} are still very popular, with research in new sub-domains emerging. In particular, several studies \cite{Kapovciute2019, Cresci2019, Guo2019, Xing2020, Chen2020, Mishev2020} focus on the Finance domain. The authors in \cite{Xing2020} identify common error patterns that result in financial sentiment analysis to fail, namely, irrealis mood, rhetoric, dependent opinion, unspecified aspects, unrecognised words, and external reference. On the other hand, in \cite{Mishev2020} Mishev et al. evaluate sentiment analysis studies in the Finance domain by starting from lexicon-based approaches and finishes with the ones that use Transformers, such as the Bidirectional Enconder Representations from Transformers (BERT) \cite{Devlin2018} and the Robustly optimised BERT approach (RoBERTa) \cite{Liu2019}. 

The ongoing coronavirus disease (COVID-19) global pandemic has led to a rise in SOM studies analysing social opinions in terms of different dimensions, such as sentiment polarity. The work in \cite{Muller2020} released a COVID-19 Transformer-based model that was pre-trained on multiple datasets of tweets from Twitter. These datasets contained tweets on various topics, such as vaccine sentiment and maternal vaccine stance, and used other well known datasets, such as SemEval 2016 - Task 4 which was previously discussed in Section \ref{ssec_social_datasets}. This model was pre-trained to carry out sentiment analysis on tweets written in other languages, such as Arabizi -- a written form of spoken Arabic that relies on Latin characters and digits \cite{Baert2020}. On the other hand, Kruspe et al. \cite{Kruspe2020} presented sentiment analysis results of 4.6 million European tweets for the initial period of COVID-19 (December 2019 till April 2020), which results were aggregated by country (Italy, Spain, France, Germany, United Kingdom) and averaged over time. An ANN was trained to carry out sentiment analysis, which model was compared with several pre-trained models, such as BERT which is trained on BookCorpus and English Wikipedia data \cite{Devlin2018}, a multilingual version of BERT trained on COVID-19 tweets \cite{Muller2020}, and the Embeddings from Language Models (ELMO) trained on the 1 Billion Word Benchmark dataset.

In terms of NLP tools, Hugging Face\footnote{https://huggingface.co/} provides a state-of-the-art Transformer library for Pytorch and TensorFlow 2.0\footnote{https://huggingface.co/transformers/}. Therefore, it provides general-purpose architectures, such as BERT, GPT-2 \cite{Radford2019}, RoBERTa, cross-lingual language model (XLM) \cite{Lample2019}, DistilBert \cite{Sanh2019}, and XLNET \cite{Yang2019} for NLP tasks (like sentiment analysis), where over 32+ pre-trained models are available in 100+ languages. Similarly, TensorFlow Hub\footnote{https://www.tensorflow.org/hub} provides a repository of trained machine learning models, with a variety of them using the Transformer architecture\footnote{\url{https://tfhub.dev/google/collections/transformer_encoders_text}}, such as BERT. 

The carbon footprint for training new deep learning models should always be taken in consideration especially if a large number of Central Processing Units (CPUs), Graphical Processing Units (GPUs), or Tensor Processing Units (TPUs) are needed. This in turn increases the related costs for model training, which is becoming very expensive and is expected to keep increasing in the future. In \cite{Strubell2019}, Strubell et al. mention that such costs amount to both the financial aspect in terms of hardware and electricity or cloud compute time, and the environmental aspect in terms of carbon footprint needed to fuel modern tensor processing hardware. Therefore, researchers should report the training time and computational resources needed in their published work, and they should prioritise computationally efficient algorithms and hardware that need less energy.
\subsection{Conclusion}
\label{ssec_conc}
The main aim of this systematic review is to provide in-depth analysis and insights on the most prominent technical aspects, dimensions and application areas of SOM. The target audience of this comprehensive review is three fold: 
\begin{itemize}
	\item Early-Stage Researchers who are interested in working within this evolving research field of study and/or are looking for an overview of this field; 
	\item Experienced Researchers already working in SOM who would like to progress further on the technical side of their work and/or looking for weaknesses in the the field of SOM;
	\item Early-Stage and/or Experienced Researchers who are looking into applying SOM/their SOM work in a real-world application area.
\end{itemize}

The identification of the current literature gaps within the SOM field of study is one of the main contributions of this systematic review. An overview below provides a pathway to future research and development work: 

\begin{itemize}
\item \textbf{Social Media Platforms}: Most studies focus on data gathered from one social media platform, with Twitter being the most popular followed by Sina Weibo for Chinese targeted studies. It is encouraged to possibly explore multi-source information by using other platforms, thus use data from multiple data sources, subject to any existing API limitations\footnote{Due to GDPR, API coverage in terms of which data can be accessed is being tightened in terms of control, which can be a major issue faced by researchers.}. 
This shall increase the variety and volume of data (two of the V's of Big Data) used for evaluation purposes, thus ensuring that results provide more reflective picture of society in terms of opinions. The use of multiple data sources for studies focusing on the same real-world application areas are also beneficial for comparison purposes and identification of any potential common traits, patterns and/or results. Mining opinions from multiple sources of information also presents several advantages, such as higher authenticity, reduced ambiguity and greater availability \cite{Balazs2016}.
\item \textbf{Techniques}: The use of Deep Learning, Statistical, Probabilistic, Ontology and Graph-based approaches should be further explored both as standalone and/or part of hybrid techniques, due to their potential and accessibility. In particular, Deep Learning capabilities has made several applications feasible, whereas Ontologies and Graph Mining enable fine-grained opinion mining and the identification of relationships between opinions and their enablers (person, organisation, etc.). Moreover, ensemble Machine Learning and Deep Learning methods and fine-tuned Transformed-based models are still under-explored. In such a case, researchers should be attentive to the carbon footprint needed to train neural network models for NLP. 
\item \textbf{Social Datasets}: The majority of available datasets are either English or Chinese specific. This domain needs further social datasets published under a common open license for use by the public domain. These should target any of the following criteria: bi-lingual/multilingual data, and/or annotations of multiple opinion dimensions within the data, e.g., sentiment polarity, emotion, sarcasm, irony, mood, etc. Both requirements are costly in terms of resources (time, funding and personnel), domain knowledge and expertise. 
\item \textbf{Language}: The majority of the studies support one language, with English and Chinese being the most popular. Studies that support two or more languages is one of the major challenges in this domain due to numerous factors, such as cultural differences and lack of language-specific resources, e.g., lexicons, datasets, tools and technologies. This domain also needs more studies that focus on code-switched languages and less-resourced languages, which shall enable the development of certain language resources needed for the respective code-switched and less-resourced languages. 
\item \textbf{Modality}: Bi-/Multi-modal SOM is another sub-domain that requires several research. Several studies cater for the text modality only, with the visual - image modality gaining more popularity. However, the visual - video and audio modalities are still in their early research phases with several aspects still unexplored. This also stems from a lack of available visual, audio and multimodal datasets. 
\item \textbf{Aspect-based SOM}: Research in this sub-domain is increasing and developing, however, it is far from the finished article, especially when applied in certain domains. Further aspect-based research is encouraged on other opinion dimensions other than sentiment polarity, such as emotions and moods, which is still unexplored. Moreover, more research is required on the use of Deep Learning approaches for such a task, which is still at an early stage.
\item \textbf{Application areas}: Most studies target Politics, Marketing \& Advertising \& Sales, Technology, Finance, Film and Healthcare. Research into other areas/sub-domains is encouraged to study and show the potential of SOM.
\item \textbf{Dimensions of SOM}: Most studies focus on subjectivity detection and sentiment analysis. The area of emotion analysis is increasing in popularity, however, sarcasm detection, irony detection and mood analysis are still in their early research phases. Moreover, from the analysis of this systematic review it is evident that there is a lack of research on any possible correlations between the different opinion dimensions, e.g., emotions and sentiment. Lastly, no studies cater for all the different SOM dimensions within their work. 
\end{itemize}

Shared evaluation tasks, such as International Workshop on Semantic Evaluation (SemEval), focused on any one of the current research gaps identified above, are very important and shall contribute to the advancement of the SOM research area. Therefore, researchers are encouraged to engage in these tasks through their participation and/or organisation of new tasks, since these shall advance the SOM research area.

In conclusion, as identified through this systematic review, a fusion of social opinions represented in multiple sources and in various media formats can potentially influence multiple application areas.

\section*{Acknowledgments}
This work is funded by the ADAPT Centre for Digital Content Technology which is funded under the SFI Research Centres Programme (Grant number 13/RC/2106).

\section{Compliance with Ethical Standards}
\begin{itemize}
	\item Conflict of Interest: The authors declare that they have no conflict of interest.
	\item Ethical approval: This article does not contain any studies with human participants or animals performed by any of the authors.
	\item Informed consent: Not applicable.
\end{itemize}

\bibliography{arxiv_2020_bibliography}

\end{document}